\documentclass{article}

\usepackage{arxiv}

\usepackage[utf8]{inputenc} 
\usepackage[T1]{fontenc}    
\usepackage{hyperref}       
\usepackage{url}            
\usepackage{booktabs}       
\usepackage{amsfonts}       
\usepackage{nicefrac}       
\usepackage{microtype}      
\usepackage{lipsum}
\usepackage{graphicx}
\graphicspath{ {./images/} }

\usepackage[table,xcdraw]{xcolor}
\usepackage{subcaption}
\usepackage{multirow}
\usepackage{pgffor}
\usepackage{arydshln}
\usepackage{colortbl}
\usepackage{xcolor}  
\definecolor{lightgray}{gray}{0.7}
\usepackage{amsmath}
\usepackage{amssymb}
\usepackage{float}

\def\MyTransformer{AquaCast}
\def\SynthNode{100}

\title{AquaCast: Urban Water Dynamics Forecasting with  Precipitation-Informed Multi-Input Transformer}

\author{
 Golnoosh Abdollahinejad \\
  EPFL and Empa\\
  Thun, Switzerland \\
  \texttt{golnoosh.abdollahinejad@epfl.ch} \\
   \And
 Saleh Baghersalimi \\
  Empa\\
  Thun, Switzerland \\
  \texttt{saleh.baghersalimi@empa.ch} \\
  \And
 Denisa-Andreea Constantinescu \\
  EPFL\\
  Lausanne, Switzerland \\
  \texttt{denisa.constantinescu@epfl.ch} \\
  \AND
  Sergey Shevchik \\
  Empa \\
  Thun, Switzerland \\
  \texttt{sergey.shevchik@empa.ch} \\
  \And
  David Atienza \\
  EPFL \\
  Lausanne, Switzerland \\
  \texttt{david.atienza@epfl.ch} \\
}

\begin{document}
\maketitle
\begin{abstract}
This work addresses the challenge of forecasting urban water dynamics by developing a multi-input, multi-output deep learning model that incorporates both endogenous variables (e.g., water height or discharge) and exogenous factors (e.g., precipitation history and forecast reports). Unlike conventional forecasting, the proposed model, \MyTransformer\, captures both inter-variable and temporal dependencies across all inputs, while focusing forecast solely on endogenous variables. Exogenous inputs are fused via an embedding layer, eliminating the need to forecast them and enabling the model to attend to their short-term influences more effectively.
We evaluate our approach on the LausanneCity dataset, which includes measurements from four urban drainage sensors, and demonstrate state-of-the-art performance when using only endogenous variables. Performance also improves with the inclusion of exogenous variables and forecast reports. To assess generalization and scalability, we additionally test the model on three large-scale synthesized datasets, generated from MeteoSwiss records, the Lorenz Attractors model, and the Random Fields model, each representing a different level of temporal complexity across 100 nodes. The results confirm that our model consistently outperforms existing baselines and maintains a robust and accurate forecast across both real and synthetic datasets.
\end{abstract}

\section{Introduction}
\label{sec:intro}
Understanding urban hydrological processes is essential to build sustainable and resilient cities. Urban drainage systems are critical for protecting public health, ensuring sanitation, protecting groundwater, and preventing floods in densely populated areas \cite{blumensaatuwo}. With the steady growth of urban populations, it becomes increasingly important to assess the potential impacts of future climatic conditions on urban environments \cite{burgstall2021urban}. Large-scale monitoring of urban water systems plays a vital role in managing this growth effectively.
However, challenges remain, including insecure environments and the need for specialized tools, which complicate and increase the costs of data collection in urban water systems~\cite{blumensaatuwo}. Tackling these challenges is essential for improving the efficiency and reliability of forecasting solutions. Time series forecasting provides a powerful tool for predicting the future status of urban water systems. By using such methods, cities can adapt their water management strategies to optimize operations in the face of climate change, ensuring sustainability. Forecasting hydrological systems also helps prevent overflows, optimize drinking water supply, manage wastewater systems, and detect anomalies. UrbanTwin~\cite{UrbanTwin} demonstrates the value of forecasting within urban digital twin frameworks for real-time decision-making.
 
The City of Lausanne in Switzerland serves as a full-scale testbed in which we assess and push the limits of the forecasting of complex urban wastewater networks. We developed \textbf{\MyTransformer}, a data-driven methodology to design AI models that anticipate rainfall-induced surges. This enables operators to (i) pre-emptively allocate or release temporary storage, (ii) keep stormwater and sewage streams separate, and (iii) shield the urban drainage systems from hydraulic or contaminant shocks during extreme events. These long-horizon forecasts further steer maintenance schedules, pollution mitigation actions, and control strategies in pipe sections that lack dense sensing or detailed hydraulic models. By delivering earlier and more accurate forecasts, our approach reduces the probability of untreated overflows into Lac Léman and nearby waterways, preserving aquatic ecosystems, curbing eutrophication, and protecting public health. 
To probe robustness and scalability toward larger metropolitan areas, we augment these field observations with three physically informed synthetic datasets that extend coverage to the wider Lausanne catchment with higher complexity to preserve robustness and consider more distributed sensors to scale up the time-series forecasting. Although validated in Lausanne, the methodology is readily transferable to other cities.

\subsection{Background}
Classical forecasting methods, such as ARMA and ARIMA \cite{box2015time} provide theoretical guarantees; however, they often struggle to capture complex interactions in real-world data, leading to suboptimal performance on multi-variate time series. In contrast, deep neural networks have demonstrated robust performance and provided promising results across a variety of real-world applications \cite{oreshkin2019n, zhou2021informer}.
Recent time series forecasting models have shown a significant capability in detecting hierarchical patterns \cite{petropoulos2022forecasting}, particularly with the advancements in deep learning (DL) models.
Forecasting time series using DL models started with multilayer perceptron (MLP) \cite{dudek2013forecasting} and recurrent networks, such as Recurrent Neural Networks (RNN) \cite{lai2018modeling}, Generalized Regression Neural Networks (GRNN) \cite{dudek2015generalized}, and Long Short-Term Memory Networks (LSTM) \cite{zheng2017electric}. All recurrent networks are based on propagation over time, making training inefficient for longer forecasts \cite{279181}. This inefficiency arises from several key issues: vanishing gradients, where early inputs lose influence as gradients shrink over time \cite{279181}; exploding gradients, which cause unstable updates; limited memory capacity, restricting long-term information retention; a bias toward recent inputs, leading to poor modeling of distant dependencies; and inherently sequential processing, which hinders efficient learning over long sequences compared to parallelable models \cite{10.1162/neco.1997.9.8.1735}.
Recently, Transformers based on attention mechanism \cite{vaswani2017attention}, have demonstrated superior performance in long-range forecasting compared to recurrent networks \cite{zhou2021informer, zhang2023crossformer, nie2022time, liu2023itransformer}. Their advantages include shorter signal propagation paths and the absence of propagation over time \cite{vaswani2017attention, zhou2021informer}. Given the complexity of real-world scenarios like urban drainage systems which include distributed locations with variable parameters, multi-variate time series forecasting is essential. This study focuses primarily on transformer networks due to their remarkable performance in sequential forecast and their utilization of the attention mechanism.

Different multi-variate Transformer networks are based on a modification to the vanilla Transformer architecture and its components. The first class, such as \cite{wu2021autoformer, zhou2021informer}, adapts the attention module for long sequences, but a simple linear forecaster \cite{zeng2023transformers} outperforms them. The second class, such as PatchTST\cite{nie2022time}, addresses the non-stationarity of time series by employing a full channel-independent Transformer, resulting in lower mean squared error (MSE). However, this channel-independent design sacrifices interpretability among variables for robustness.
Crossformer \cite{zhang2023crossformer}, in the third class, modifies both the architecture and components to account for dependencies in both the temporal domain (cross-time) and among multiple variates (cross-variate). Despite its advanced architecture, the router mechanism in the cross-variate stage limits the access to the attention map.
In contrast, iTransformer \cite{liu2023itransformer} retains the original Transformer components but applies them to inverted dimensions with a revised architecture. This allows it to capture cross-variate dependencies in its attention maps, where each input token represents an embedded univariate time series. While iTransformer performs better than Crossformer by using an untouched Transformer, it lacks the ability to capture cross-time dependencies.

\begin{figure}
    \centering
    \includegraphics[width=1\linewidth]{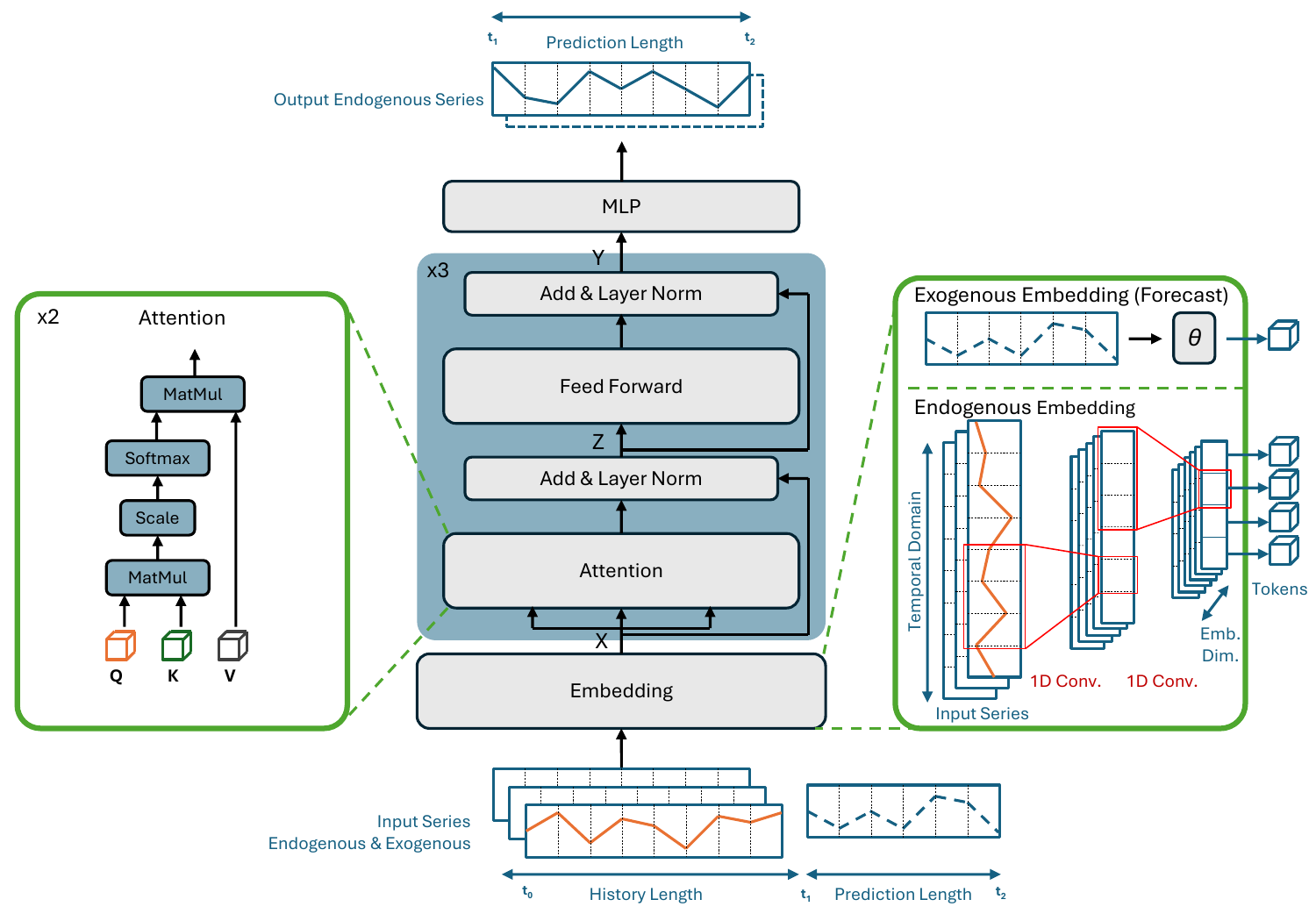}
    \caption{\MyTransformer\ network architecture: input time-series, can be history and forecast, are embedded (right), and then are fed into Attention mechanism (left). The transformer block (middle) provide the encoded tokens and then decoded using MLP layer (top) to output the forecasted series for only selected endogenous series.}
    \label{fig:MyTransformer}
\end{figure}

\subsection{Related Work}
Urban water forecasting has seen substantial advancements with the adoption of deep learning methods. Transformer-based models have shown superior forecast accuracy over traditional methods by capturing complex temporal patterns. For example, \cite{Demiray2024} applied transformer architectures to stream-flow forecasting and found them superior. 
Compared to LSTM, Gated Recurrent Unit (GRU), and Seq2Seq models, though they noted sensitivity to data preparation methods. This work asked to investigate larger scale dataset. 
Hybrid models integrating LSTM and transformers have also shown promising results. A hybrid algorithm in~\cite{Li2024} developed a rainfall-runoff forecasting model optimized with random search, which demonstrated high precision and robustness over standalone LSTM and transformer models, however, this work does not investigate increasing the accuracy by incorporating exogenous variables, such as environmental factors for flood forecasting. This work asked to investigate increasing the accuracy by incorporating exogenous variables, such as environmental factors for flood forecasting.

Similarly, researchers in~\cite{Zhou2022} combined CNN and LSTM networks with attention mechanisms, achieving superior accuracy in daily urban water demand forecasting. This attention-based CNN-LSTM framework efficiently managed temporal and multivariate dependencies by considering weather forecast information, in addition to only the daily max and min temperature and encoded weather types but not the precipitation forecast in form of hourly time-series.

Recent hybrid approaches integrating GRU, Temporal Convolutional Network(TCN), and Transformer architectures have further advanced forecasting precision. To forecast gate-front water level~\cite{Zhao2024} utilized Singular Spectrum Analysis (SSA) and Complete Ensemble Empirical Mode Decomposition with Adaptive Noise (CEEMDAN) alongside a permutation entropy algorithm to enhance the forecasting. Their GRU–TCN–Transformer coupled model significantly outperformed traditional machine learning and individual deep learning models.
Furthermore, the study does not consider the inclusion of other variables that might influence water dynamics near the sluice gate. It is necessary to increase the diversity of input data to improve the adaptability of the model to complex environmental factors and improve the comprehensiveness of the forecasts.
For water quality forecasting, ~\cite{xue2025study} also utilizes the attention-based network and adds the correlation to achieve high-precision long-term forecast of water quality data, its effectiveness for shorter-term forecasts is comparatively lower than traditional methods.
For water demand forecasting, ~\cite{lin2025multi} modify PatchTST and achieve remarkable results compared to other DL architectures. Although PatchTST is widely used in forecasting systems, which has also been used in hydrological systems too, lack of considering inter-channel dependencies makes it unable to provide information for other variables during the forecasting.

Broadly speaking, exploring the effectiveness of transformer-based networks for forecasting water-related variables~\cite{li2025review}, particularly under specific application settings such as exogenous and target variables, holds significant potential.

\subsection{Contribution}
This work investigates urban water dynamics using real-world data from the city of Lausanne alongside with the large-scale synthesized datasets. By creating a clean, structured dataset from these sources, we establish a foundation for analyzing key forecasting challenges in urban water systems. Our goal is to present, \MyTransformer\, a deep learning model suited to this domain, benchmark their performance, and address common real-world challenges. This contributes to developing scalable forecasting solutions for complex urban environments. 
Unlike traditional multi-to-multi forecasting methods, our setup, \MyTransformer, can focus on predicting a single fixed output variable or performing multi-variate forecasting of selected endogenous variables by considering both the history of exogenous variables and forecast information. This enables a fair evaluation of the improvements achieved through shared information among multiple input series.
The following points summarize our contributions:

    \begin{itemize}
    \item To understand and address the forecast challenges in a newly instrumented urban catchment, we perform an in-depth analysis of real-life data from part of the urban drainage system in Lausanne. This involves comprehensive data exploration, cleaning, and the creation of a robust, structured dataset tailored for short-term water dynamics forecasting.
    \item To empower city officials with a holistic forecasting system that integrates endogenous and exogenous time series, we designed a transformer-based network, \MyTransformer, capable of incorporating exogenous variables, such as precipitation history and forecast, along with endogenous water drainage measurements on water height and discharge.
    \item To enhance predictive capabilities in real world conditions, we modify our architecture to incorporate forecast input, specifically integrating precipitation forecasts into the forward pass of \MyTransformer.
    \item To balance short-term responsiveness with long-term reliability, we systematically examine the trade-off between forecast accuracy and forecast horizon, helping identify \MyTransformer\ configurations best suited for operational deployment.
    \item To ensure applicability across varying urban contexts and scales, we evaluated the scalability of our approach by generating and utilizing synthetic datasets with varying degrees of structural complexity, simulating larger-scale urban drainage systems.
    \end{itemize}
The remainder of this paper is organized as follows: Section~\ref{sec:method} presents the proposed methodology and forecasting framework. Section~\ref{sec:setup} describes the experimental setup, including datasets and evaluation metrics. Section~\ref{sec:result} discusses the results and performance analysis. Finally, Sections~\ref{sec:discussion} and ~\ref{sec:conc} discuss and conclude the paper.

\section{Methodology}
\label{sec:method}



\subsection{Model Architecture}


Unlike traditional multi-to-multi approaches, \MyTransformer\ supports both single-variable and multi-variable forecasting by leveraging exogenous variable history and forecast reports. This enables fair evaluation of information sharing across input time series.. The model begins with a unified embedding layer for both input types, followed by a standard Transformer encoder with multi-head self-attention.

\textbf{Embedding.} This layer consists of two sections. For history data consists of two stacked one-dimensional convolution (Conv1D) layers, as illustrated on the right side of Figure \ref{fig:MyTransformer} that project from $\mathbb{R}^{\text{History\_Length}}$ to $\mathbb{R}^D$. The first Conv1D layer extracts coarse-grained temporal features by summarizing broad segments of the input sequence. It also serves as a patching mechanism, inspired by PatchTST~\cite{nie2022time}, but applied jointly across all input series rather than in a channel-independent approach. This ensures that each temporal token integrates information from the full multivariate context. The second Conv1D layer builds on the output of the first, enabling the model to learn higher-level abstractions from the previously extracted features. A non-linear activation function, specifically ReLU, is applied between the two layers to enhance representational capacity. This hierarchical encoding allows the attention block to capture inter-temporal dependencies while considering inter-variate dependencies, reducing the need for a router and a separate cross-variable attention mechanism as in CrossFormer~\cite{zhang2023crossformer}. The embedding layer can also consider the forecast info by applying a trainable linear projector from $\mathbb{R}^{\text{Forecast\_Length}}$ to $\mathbb{R}^D$. This single token that resembles future information will be treated as an extra input token with the same model dimension without increasing the trainable parameters of attention block. Having a single embedded token for each forecasted variable allows the model to handle different lengths for history and forecast sequences. It also supports the use of forecasted variables with a different temporal resolution than historical data, making the model more flexible and suitable for a wider range of forecasting tasks.


\textbf{Transformer Encoder.} The attention mechanism within the Transformer is based on the Query-Key-Value paradigm. The right side of Figure~\ref{fig:MyTransformer} visualizes this block. Based on Equation \ref{eq:transformer_z} each input token is projected into query, key, and value vectors. The matrices $W_Q$, $W_K$, and $W_V$ represent the learnable weights, respectively.
The attention weights are computed by evaluating the similarity between the query and all keys, resulting in a weighted sum of the values. This operation enables the model to focus selectively on the most relevant parts of the sequence. Equation~\ref{eq:attention} provides the details of this process.

Following the attention block, a feed-forward neural network is applied on output tokens. This network further processes the refined token representations to extract higher-level features. Additional components such as layer normalization and residual connections are included to stabilize training and enhance convergence. This part is formalized in Equation~\ref{eq:transformer_y}.

\begin{equation}
\text{Attention}(Q, K, V) = \text{softmax}\left(\frac{Q K^T}{\sqrt{d_k}}\right)V
\label{eq:attention}
\end{equation}

\begin{equation}
Z = \text{LayerNorm}(X + \text{Attention}(X W_Q, X W_K, X W_V))
\label{eq:transformer_z}
\end{equation}

\begin{equation}
Y = \text{LayerNorm}(Z + \text{FeedForward}(Z))
\label{eq:transformer_y}
\end{equation}

By stacking these layers, the Transformer builds a hierarchical representation of the input sequence. Each layer captures increasingly abstract patterns over time. This allows the model to learn both short-term and long-term dependencies in the data. As a result, the Transformer is able to model complex temporal relationships effectively.

\textbf{Decoder.} After the Transformer layers, the full set of output tokens is passed to a single-layer perceptron. This perceptron performs regression over the entire output, using information from all tokens. It produces the forecasted values for each time step in the target series based on the combined sequence representation.

\begin{table}
\renewcommand{\arraystretch}{1.3}
\centering
\resizebox{\textwidth}{!}{%
\begin{tabular}{|
>{\columncolor[HTML]{FFFFFF}}c |
>{\columncolor[HTML]{FFFFFF}}c |
>{\columncolor[HTML]{FFFFFF}}c |
>{\columncolor[HTML]{FFFFFF}}c |
>{\columncolor[HTML]{FFFFFF}}c |
>{\columncolor[HTML]{FFFFFF}}c |
>{\columncolor[HTML]{FFFFFF}}c |}
\hline
\textbf{\#} & \textbf{Sensor} & \textbf{Sensor Type} & \textbf{Metric} & \textbf{Resolution} & \textbf{Max Height (mm)} & \textbf{Discharge Estimation} \\ \hline
1 & A & IJINUS LNU06V3/V4 & Water Height & 1 min & $\sim$1000 & Rating curve (Manning) \\ \hline
2 & B & IJINUS LNU06V3/V4 & Water Height & 1 min & $\sim$600 & Rating curve (Manning) \\ \hline
3 & C & Doppler sensor & Discharge, m³/s & 15 min & 1500 & Direct (Doppler), $\sim$15\% uncertainty \\ \hline
4 & D & Doppler sensor & Discharge, m³/s & 15 min & 2100 & Direct (Doppler), $\sim$15\% uncertainty \\ \hline
\end{tabular}
}
\caption{Sensor Overview, Measurement Types, and Resolution}
\label{table:sensor_overview}
\end{table}

\section{Experimental Setup}
\label{sec:setup}
We trained all models and configurations using the Adam optimizer and the mean squared error (MSE) as the loss function. Training was performed on an NVIDIA GeForce RTX 4090 GPU for a maximum of 100 epochs. To prevent overfitting, an early stopping strategy with a patience of 10 epochs was employed, terminating training if no improvement was observed on the validation set. The number of different layers of each model is set by hyperparameter tuning to have the most accurate results while keeping the size of the model as small as possible.

In the following subsections, we will present the datasets utilized in this study. The first set is real-world data acquired from the LausanneCity drainage system. To scale up the dataset and study catastrophic cases, we introduce 3 synthesized datasets on LausanneCity terrain and wastewater graph with different levels of complexity. After structuring the dataset version, we will review the metrics used for evaluation.

\subsection{LausanneCity Dataset} 

\begin{figure}
    \centering
    \includegraphics[width=0.85\linewidth]{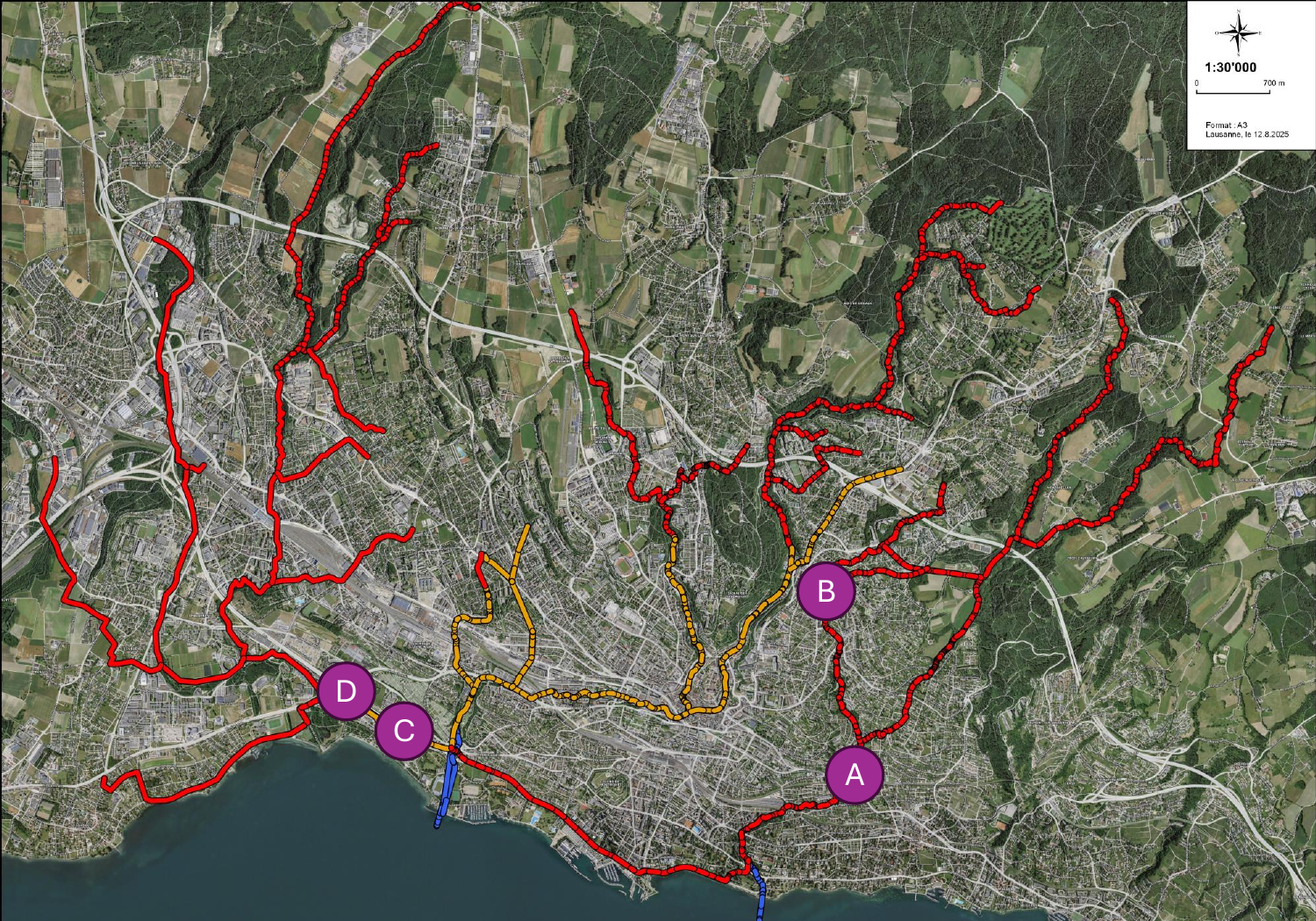}
    \caption{Sensors locations on Lausanne city map and connections of waste water pipes. Selected sensors are highlighted in purple. sensors \textit{C} and \textit{D} represent together the total getting the wastewater treatment plant (WWTP)}
    \label{fig:city-map}
\end{figure}

The main dataset used in this study includes records of the water levels in drainage system of the Ouchy area in Lasuanne city. It consists of water dynamics measurements from four sensor in four locations, making it well suited for testing data-driven modeling approaches for water feature forecasting. Two recorded water heights at 1-minute intervals and two water discharge at 15-minute intervals, as described in Table~\ref{table:sensor_overview}. Figure ~\ref{fig:city-map} shows the sensor locations and their connections through the urban drainage system. Selected sensors are highlighted in purple. Although Sensor 3.~\textit{C} and Sensor 4.~\textit{D} are not connected directly, together, they  represent the total Lausanne discharge getting the wastewater treatment plant (WWTP). According to \textit{Service de l’Eau}~\cite{LausanneCity}, Lausanne is a steep city, which results in very short lag times between rainfall events and their hydrological response, making accurate forecasting especially valuable; in contrast, flat cities exhibit different drainage dynamics and longer response times.

All water level measurements were collected using IJINUS LNU06V3 or LNU06V4 ultrasonic level sensors~\cite{sensors}.
These sensors are autonomous battery-powered devices designed for use in drainage systems. They measure water height using aerial ultrasonic waves and support flow transformation using predefined rating curves. Maximum water heights vary by site and correspond to the pipe dimensions: around 1000mm for sensor A, 600mm for Vuachère Vallon.
The discharge at these sites is estimated using a Manning-Strickler-based rating curve~\cite{chartrand2018morphodynamics}.

In contrast, the two sensors labeled \textit{C} and \textit{D} are equipped with sensors that measure both water level and velocity using Doppler technology. These enable direct discharge estimation with lower uncertainty. The measurement range for sensor \textit{C} is 0 to 1500mm, and for sensor \textit{D} is 0 to 2100mm.


\begin{table}
\renewcommand{\arraystretch}{1.3}
\centering
\resizebox{0.4\columnwidth}{!}{%
\begin{tabular}{|
>{\columncolor[HTML]{FFFFFF}}c |
>{\columncolor[HTML]{FFFFFF}}c |
>{\columncolor[HTML]{FFFFFF}}c |}
\hline
Sensor & ADF Statistic~\cite{dickey1979distribution} & p-value \\ \hline
A & -15.180 & $6.22 \times 10^{-28}$ \\ \hline
B & -12.453 & $3.54 \times 10^{-23}$ \\ \hline
C & -14.282 & $1.32 \times 10^{-26}$ \\ \hline
D & -12.916 & $3.97 \times 10^{-24}$ \\ \hline
\end{tabular}
}
\caption{Augmented Dickey-Fuller (ADF) test results for water features time-series and related p-value for null test in which we consider the time-series to be stationary if p-value is less than 0.05. The results show all time-series are stationary.}
\label{table:stationarity}
\end{table}

\begin{figure}
    \centering
    \includegraphics[width=\linewidth]{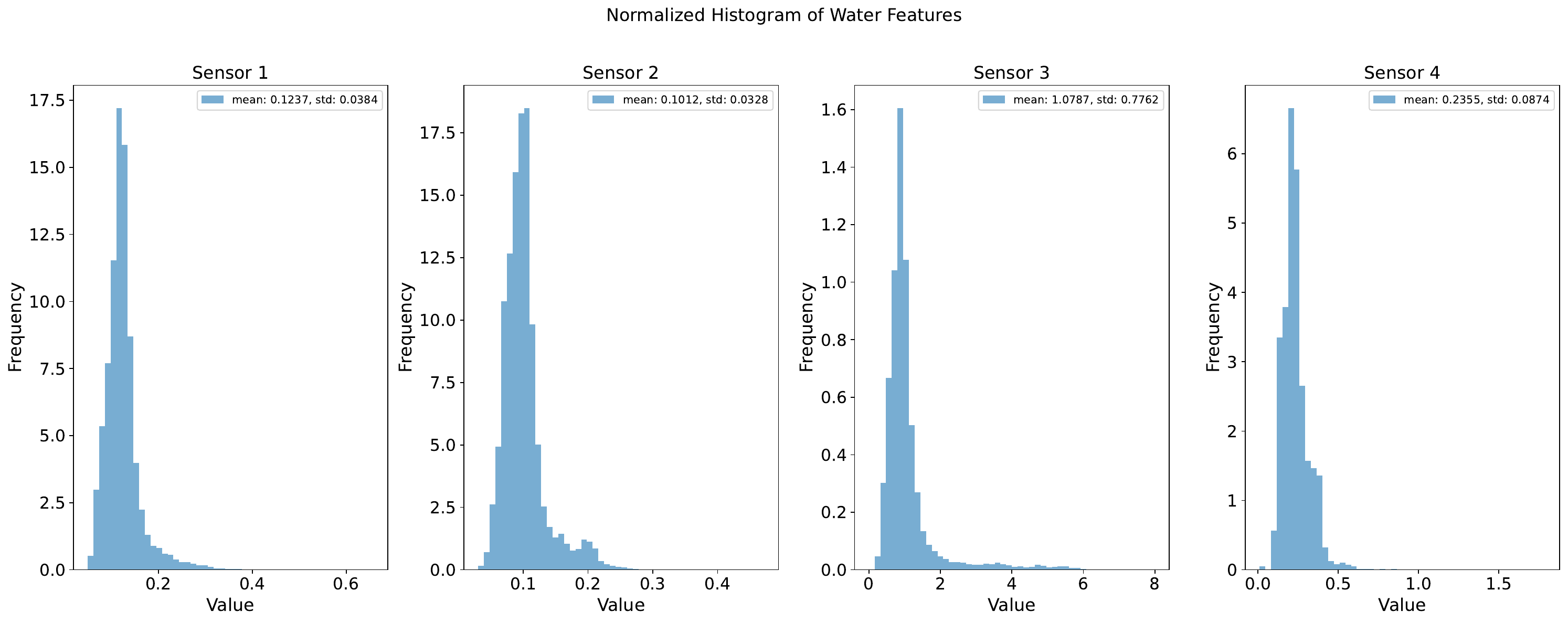}
    \caption{Normalized histogram of 4 water features with 50 bins. The mean and std reported in each legend is for raw time-series.}
    \label{fig:histogram_water}
\end{figure}

\begin{figure}
    \centering
    \includegraphics[width=0.7\linewidth]{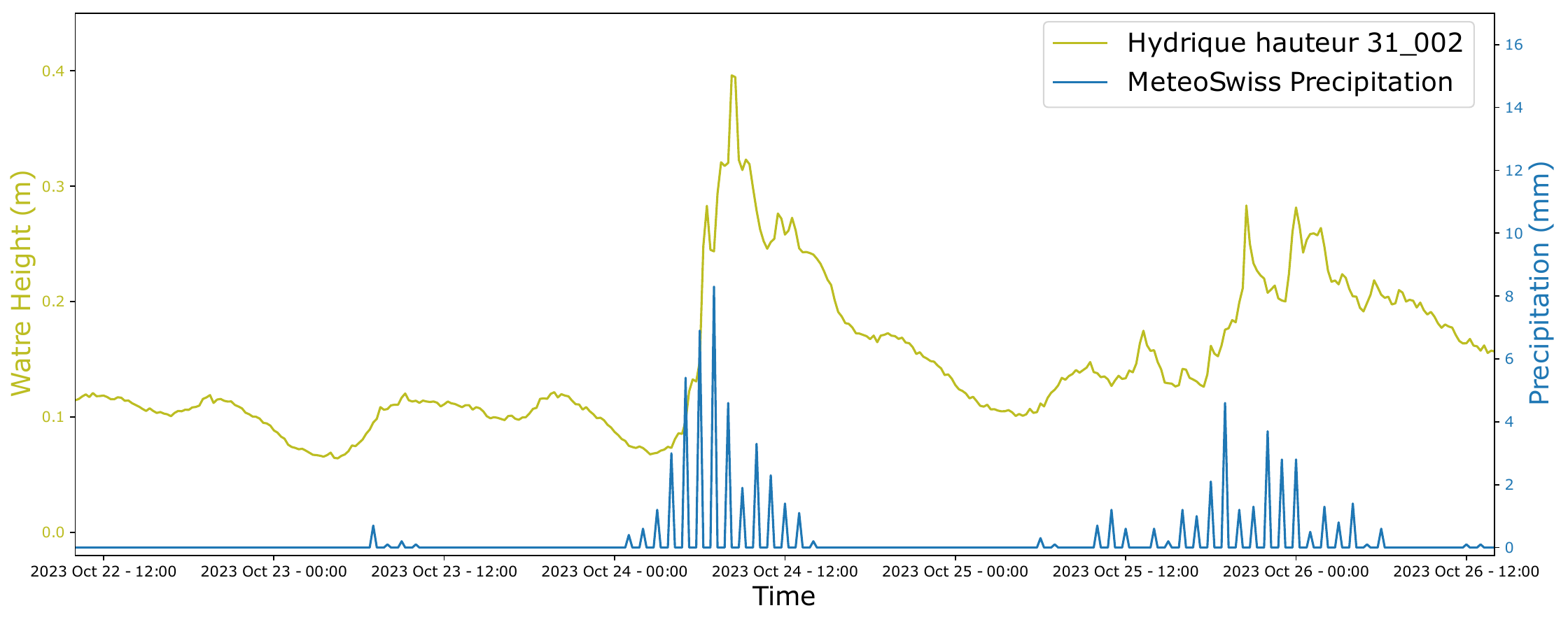}
    \caption{Plot of water height parameter (in green) and corresponding rain affect (in blue) in three days in October 2023}
    \label{fig:rain-affect}
\end{figure}

\textbf{Preprocessing} -- To prepare the dataset, 
periods where the water height remained constant for one hour - equivalent to four time-steps at final 15 minutes resolution - or longer were flagged as missing samples.
Third-order spline interpolation was applied to fill the missing time steps. After interpolation, the 1-minute resolution time-series were downsampled to 15-minute intervals to match the lower-resolution sensors and create a uniform dataset.

After preprocessing, the final dataset included four sensors with 49,673 time steps each, covering 517 days from July 2023 to early December 2024. At the final step, the dataset is split into training, validation, and test sets with 362, 52, and 103 days, respectively, in a ratio of 70\%, 10\%, and 20\%. All time-series samples are independently standardized to a zero mean and a variance of 1 based on the training set statistics before being fed into the network and are converted back to the original range after the output. 

To explore the distribution of water height and discharge measurements, normalized histograms were plotted for each sensor. As shown in Figure~\ref{fig:histogram_water}, each histogram is scaled such that the total area under the curve (AUC) equals one, allowing direct comparison across sensors with different measurements dynamic range. The distributions from the ultrasonic water level sensors, sensor 1.\textit{A} and sensor 2.\textit{B}, are both narrow and concentrated toward lower values, with means of 0.1236 and 0.1012, and standard deviations of 0.0384 and 0.0328 respectively. In contrast, the Doppler-based discharge sensors show more variability. Sensor 3.\textit{C} displays a broader and more right-skewed distribution, with a mean of 1.0784 and standard deviation of 0.7755. Sensor 4.\textit{D}, while also right-skewed, is more compact, with a mean of 0.2355 and standard deviation of 0.0874. These distinct distribution shapes reflect differing hydraulic dynamics across the monitored sites and highlight the variability captured in the dataset.

To study the stationarity of the water height time-series, we applied the Augmented Dickey-Fuller test (ADF)~\cite{dickey1979distribution} to each time series. The results are reported in Table~\ref{table:stationarity}. All four time-series showed p-values well below the 0.05 significance threshold. This suggests that the statistical properties of the time series, such as mean and variance, remain constant over time, which is beneficial for training time-series forecasting models without the need for additional differencing or de-trending procedures.

\textbf{Precipitation Records and Rain Effect} -- Alongside the water dynamic time-series, hourly precipitation records (in millimeters) from the MeteoSwiss~\cite{meteoswiss} station in Lausanne were also fed as one input. To align them with the 15-minute resolution of the water time series, the precipitation data were up-sampled by inserting zeros between recorded values.

Figure~\ref{fig:rain-affect} shows an example of the water height and precipitation data plotted together. As seen in the figure, rainfall events cause noticeable deviations from the regular periodic behavior of the water height time-series, with the degree of disruption depending on the rainfall intensity. In this work, precipitation is treated as an exogenous input to the model—used to forecast water dynamics, but not predicted itself.

In this system, each water height or discharge value is affected by two measurable factors: 
1.- A dynamic amount of water flows through the tubes.
2.- Precipitation input, also called rain in this work, is an exogenous time-series.
In the UrbanTwin framework, where spatially distributed sensors collect data from multiple locations, \MyTransformer\ automatically infers the underlying inter‐variable dependencies. In particular, for each target sensor, the model learns which other nodes applies an effect on its outputs, effectively identifying and weighting the most relevant spatial relationships.
This cycles of drainage system and how each node and its parameters affecting each other are all endogenous. In contrast, in case of precipitation inputs presence (non-zero values), make the forecast fails utilizing only endogenous information.

\subsection{Synthesized Dataset}
\label{sec:synth}
\begin{figure}
    \centering
    \begin{subfigure}{0.32\linewidth}
        \centering
        \includegraphics[width=\linewidth]{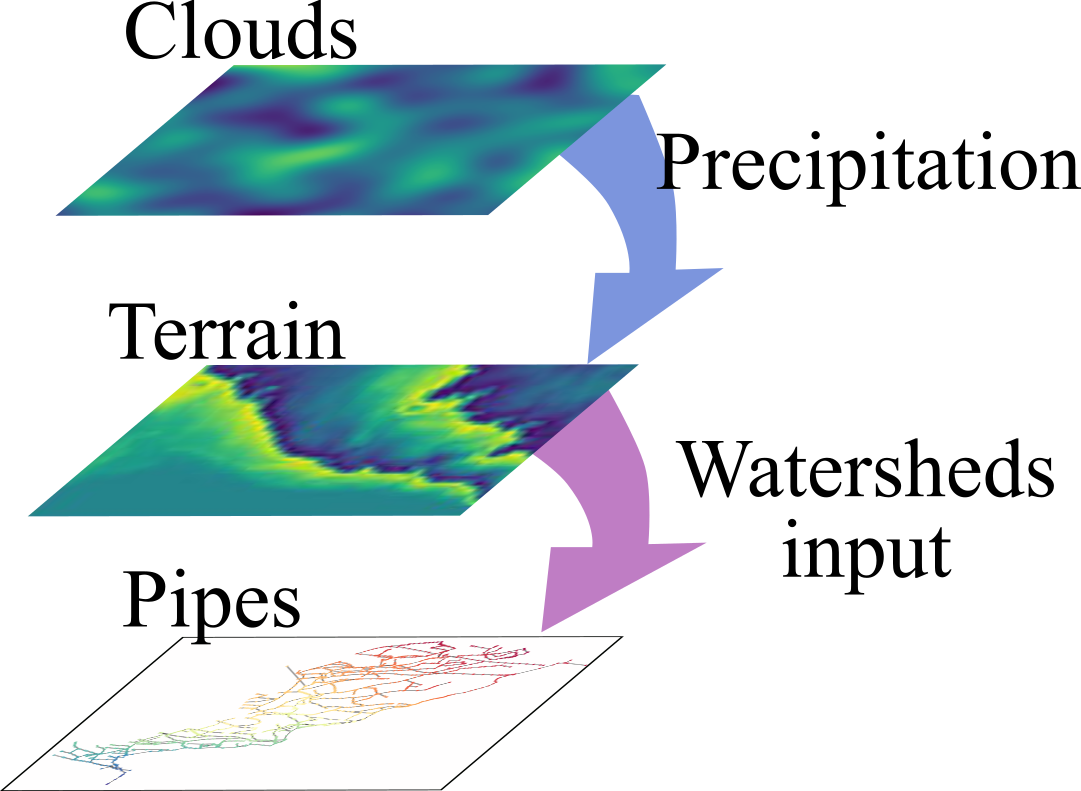}
        \caption{Precipitation Elements}
        \label{fig:synth:e1}
    \end{subfigure}
    \begin{subfigure}{0.32\linewidth}
        \centering
        \includegraphics[width=\linewidth]{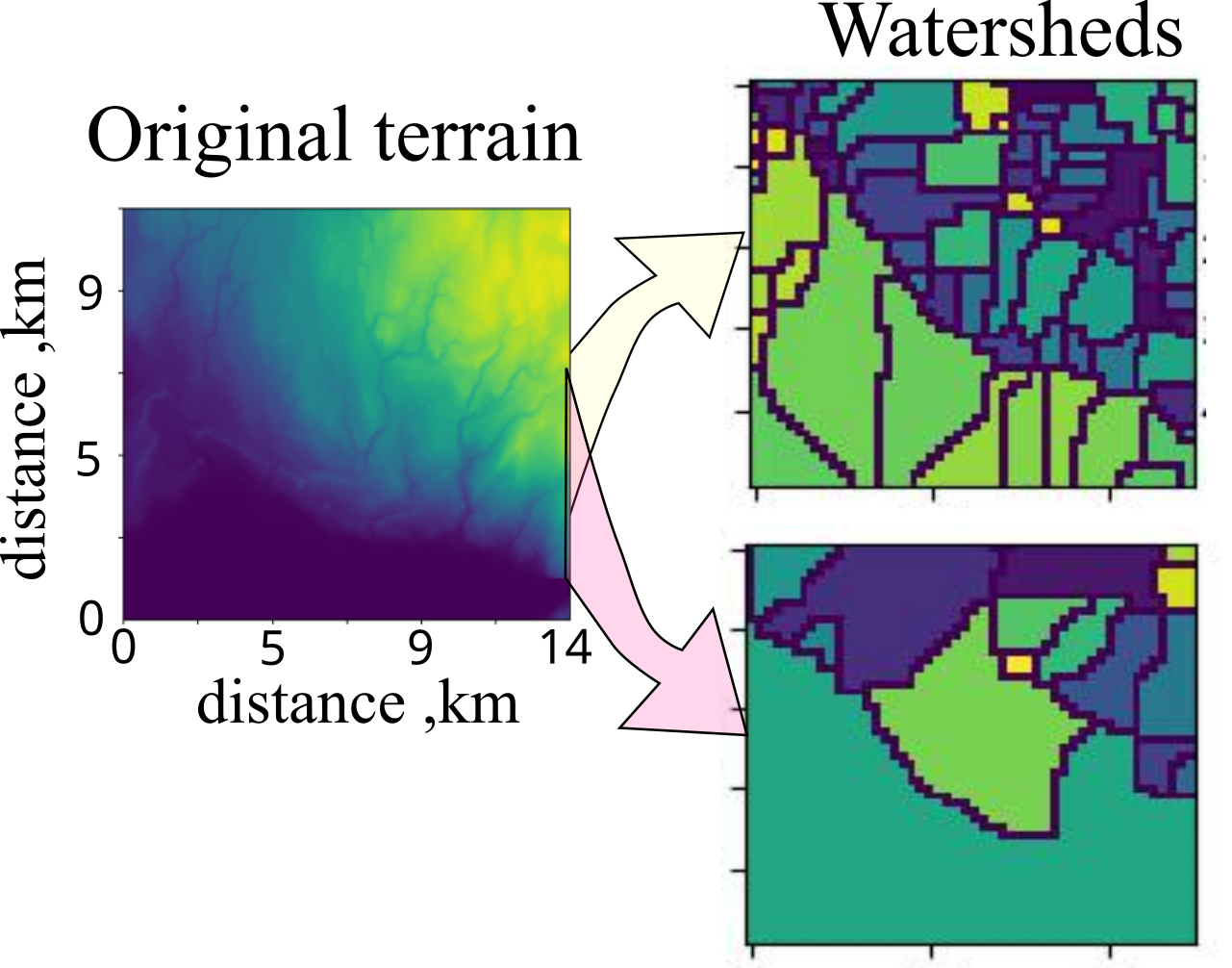}
        \caption{Terrain Elements}
        \label{fig:synth:e2}
    \end{subfigure}
    \begin{subfigure}{0.32\linewidth}
        \centering
        \includegraphics[width=\linewidth]{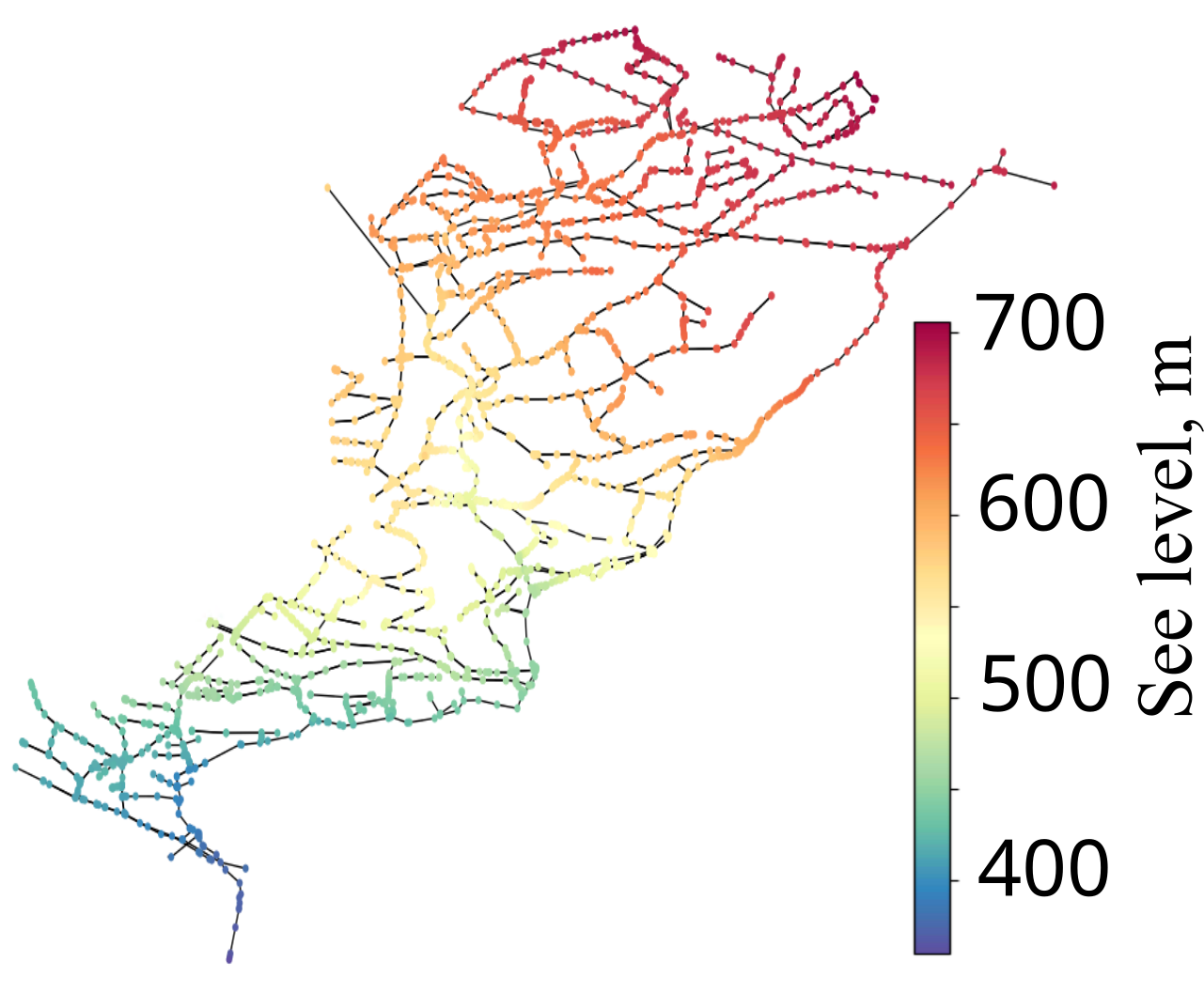}
        \caption{Pipes Networks Elements}
        \label{fig:synth:e3}
    \end{subfigure}
    \caption{The schematics of the differential model for water propagation dynamics in pipe network: a) layers of the model, b) the example of the clouds spatial distribution, c) example of the Lausanne terrain map before (top) and after (bottom) watershed segmentation, d) water pipe network, Lausanne city, including approximately 3500 nodes.}
    \label{fig:synth}
\end{figure}

Synthetic datasets are generated by a specially developed continuous differential model (CDM) of water propagation inside pipe networks. We have to notify that our CDM does not intend to precisely reproduce all the phenomena behind. However, approximating the real urban systems to some extent, this model allows the generation of water dynamics timeseries with higher complexity. Those are used  for robustness tests of \MyTransformer\ and check of its functionality in in more complex scenarios. CDM includes three main elements that are linked sequentially, while the general scheme of the model can be seen in Figure~\ref{fig:synth:e1}. The cloud element is introduced first to generate precipitations that are the input for CDM. The clouds generation is carried out in 2D spatial domain, where the local precipitation intensity is proportional to the clouds local density. The continuous shift of the clouds in spatial domain provides a sequence of unique precipitation patterns and, consequently, varying  the spatial water distribution in time. Three available clouds generation methods are introduced in our study, namely: i) real meteorological records of clouds, ii) clouds, generated with accordance to any user given deterministic law, and iii) clouds, generated with accordance to any user given stochastic law. In particular, method (i) utilizes the precipitation records from Federal Office of Meteorology and Climatology  (MeteoSwiss, Switzerland)~\cite{meteoswiss}. The time series used include 10000 discrete measurements.  Method (ii) is based on classical Lorenz attractor model ~\cite{Lorenz}.  Method (iii) utilizes random fields based on Gaussian distributions, according to approach of Müller et al. \cite{muller2022gstools}. The time series of spatial water distributions are further  passed to terrain element. In this element the input momentary water portions are redistributed with accordance to the user defined terrain topology, which conditions the further water dynamics. The terrain in this element is segmented into separated watershed regions with no water exchange  between those. The watershed segmentation is carried out using a classical approach from image processing~\cite{gonzalez2009digital}, while different detail-level of segmentation is accessible through the user defined settings. Some examples of different detail-level of watershed segmentation can be seen in Figure~\ref{fig:synth:e2}. The higher and lower number of watersheds effect more or less complex water dynamics inside CDM, respectively. Each watershed includes a unique water accumulation and propagation transfer function, dependent on the terrain profile of this specific watershed. We have to mention that the code allows to obtain terrain by i) random generation or ii) real earth maps import. In this study the real terrain map of Lausanne city was used for modeling. After the water redistribution's between different watersheds the corresponding water amounts are fed to the nodes of the underneath pipe network, included inside the next element. 

The pipe network element includes details about 2-dimension (2D) spatial locations of nodes and their pipe connections within a distributed drainage system. The alignment of 2D spatial domains of all CDM  elements, Figure~\ref{fig:synth:e2}, allows to link each node to specific watershed with respect to its spatial spatial position and provides this node water feeding from an individual watershed. The momentary input water amounts for each node are further propagated through the pipes, with all water ultimately dropping off at terminal nodes. In our CDM setup, each pipe has an individual transfer function that is defined as classical aperiodic block with delay. The delay is adjusted with respect to pipe inclination angle and length.

With all mentioned above, the CDM model reproduces a non-linear distributed system, in which even week non-linear affects between the neighbored nodes may lead to very complex dynamics on the scale of entire pipe network. To replay "close to realistic" scenarios using CDM we utilize data from Lausanne city. In particular, the real terrain of Lausanne city was involved and available at~\cite{swisstopo}. In Figure~\ref{fig:synth} the example of segmentation of real Lausanne city terrain map is carried out. The pipe network is provided by \textit{Service de l’Eau}~\cite{LausanneCity}, with the scheme in Figure~\ref{fig:synth:e3}. The dots in the scheme define the nodes, while the colors encode their height over the see level. The code of CDM also supports water dynamics modeling using several other existing popular python hydrological modeling libraries ~\cite{klise2020water, xing2020transient}, however those were not used in this study due to their lower complexity of water dynamics time series.

\begin{figure}
    \centering
    \includegraphics[width=0.5\linewidth]{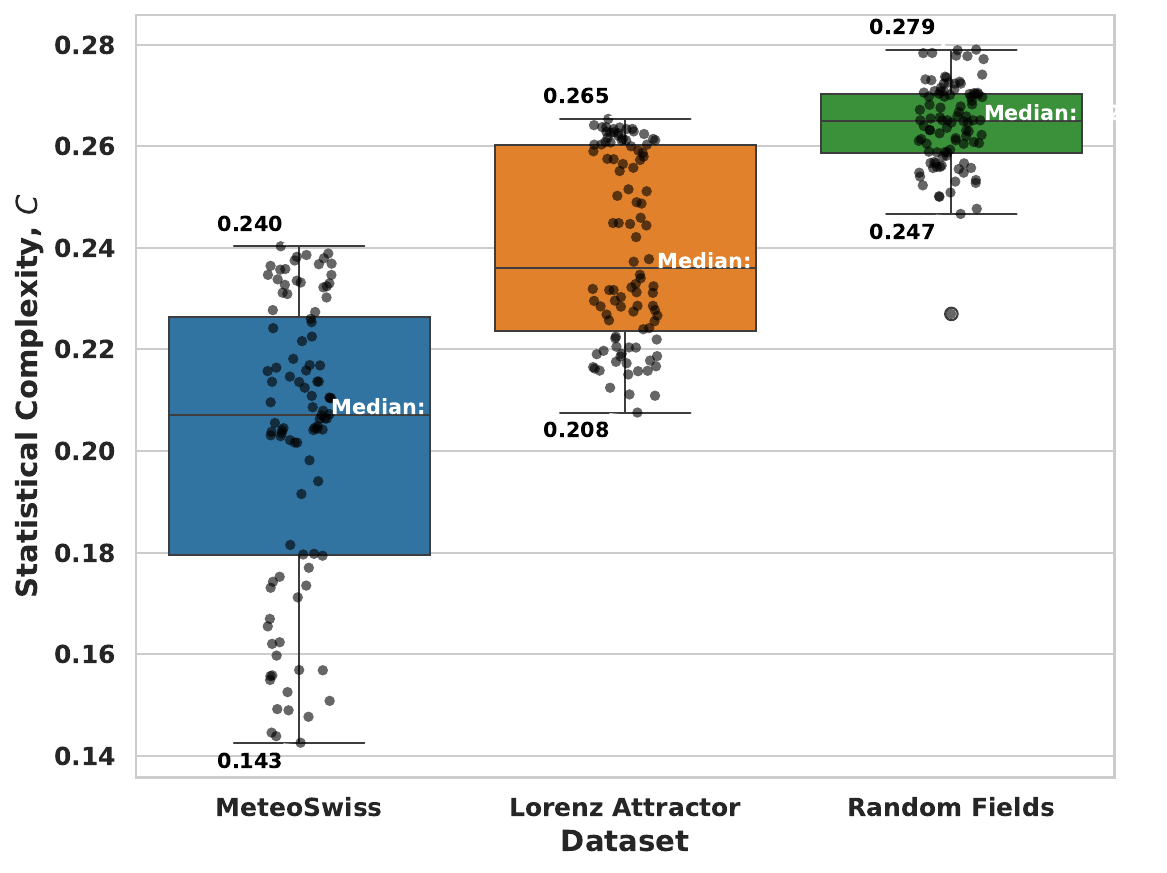}
    \caption{Statistical complexity~\ref{eq:complexity} by synthesized datasets}
    \label{fig:complexity-synth}
\end{figure}

As mentioned above, the objective of CDM is the generation of higher complexity dynamics that consequently leads to higher statistical complexity. The box and whisker plots in Figure~\ref{fig:complexity-synth} compares the distribution of statistical complexity presented in Formula~\ref{eq:complexity} across three synthesized datasets: MeteoSwiss, Lorenz Attractor, and Random Fields. The MeteoSwiss dataset shows the lowest median complexity (0.207) and a narrow interquartile range (IQR), indicating relatively simple and uniform temporal structures. The Lorenz Attractor dataset has a moderately higher median (0.236) with a wider IQR, reflecting a balance between structure and variability typical of chaotic systems. The Random Fields dataset presents the highest median complexity (0.265) and the broadest IQR, suggesting a rich diversity of temporal patterns with ordered and disordered behavior. In the box plots, the height of each box represents the IQR, and taller boxes indicate greater internal variability. This range of complexity across datasets—from low and consistent to high and diverse—provides a well-rounded basis for evaluating how forecasting models handle different levels of temporal structure and variability.

\subsection{Data Configuration}
In this work, we used two source of data. The Lausanne city records are used as real life application and synthesized data are introduced to scale up the \MyTransformer. Table~\ref{table:dataset_version} summarizes all these datasets used in this work.

\textbf{Lausanne City --}
Data configuration \textit{Lausanne\_NoRain} only consider and monitors the endogenous time-series. Dataset \textit{Lausanne\_RainHist} add history of rain to \textit{Lausanne\_NoRain} and \textit{Lausanne\_RainFull} consider also the perfect forecast from MeteoSwiss. The reason for the need to have a forecast for rain is shown in Figure~\ref{fig:rain-affect}. This plot shows the time difference between a rain peak and its effect on water features is not long enough to ensure us that the model can see it if we only use the rain history, as we did in \textit{Lausanne\_RainHist}. The idea here is that if feeding rain forecast values can help the water feature forecasting. Rain forecasts can be acquired from weather forecast models, which has good precision in Switzerland.


\textbf{Synthesized --}
For synthesized data, as explained in previous section~\ref{sec:synth} based on the source of precipitation, we construct 3 datasets. Clouds in \textit{SynthLow} is fed by MeteoSwiss records, the same as \textit{Lausanne}. Lorenz Attractor~\cite{Lorenz} and Random Fields~\cite{muller2022gstools} are the main cloud source for \textit{SynthMid} and \textit{SynthHigh} respectively. 

\begin{table}[h!]
\centering
\resizebox{0.6\columnwidth}{!}{%
\begin{tabular}{|l|c|c|l|l|}
\hline
\textbf{Data Configuration} & \textbf{Locations} & \textbf{Sensors} & \textbf{Used Rain Info} & \textbf{Rain Source} \\
\hline
Lausanne\_NoRain   & 4 & 2 & None                         & --- \\
Lausanne\_RainHist & 4 & 2 & History                 & MeteoSwiss \\
Lausanne\_RainFull & 4 & 2 & History + Forecast           & MeteoSwiss \\
\hline
SynthLow\_NoRain   & \SynthNode & 1 & None                       & --- \\
SynthLow\_RainHist & \SynthNode & 1 & History                    & MeteoSwiss \\
SynthLow\_RainFull & \SynthNode & 1 & History + Forecast         & MeteoSwiss \\
\hline
SynthMid\_NoRain   & \SynthNode & 1 & None                       & --- \\
SynthMid\_RainHist & \SynthNode & 1 & History                    & Lorenz Attractor \\
SynthMid\_RainFull & \SynthNode & 1 & History + Forecast         & Lorenz Attractor \\
\hline
SynthHigh\_NoRain   & \SynthNode & 1 & None                       &  --- \\
SynthHigh\_RainHist & \SynthNode & 1 & History                    & Random Fields \\
SynthHigh\_RainFull & \SynthNode & 1 & History + Forecast         & Random Fields \\
\hline
\end{tabular}}
\caption{Overview of LausanneCity and Synthesized datasets (MeteoSwiss~\cite{meteoswiss}, Lorenz Attractor~\cite{Lorenz}, and Random Fields~\cite{muller2022gstools}) with their rain information and sources.}
\label{table:dataset_version}
\end{table}

\subsection{Metrics}
\label{sec:metric}
For evaluating forecasting performance, we employ both point-wise and sequence-level metrics. In particular, the standard point-wise error metrics include the Mean Squared Error (MSE) and Mean Absolute Error (MAE), defined as:

\begin{equation} \text{MSE} = \frac{1}{n} \sum_{i=1}^{n} (y_i - \hat{y}_i)^2 \end{equation}

\begin{equation} \text{MAE} = \frac{1}{n} \sum_{i=1}^{n} |y_i - \hat{y}_i| \end{equation}

where $y_i$ and $\hat{y}_i$ are the ground truth and predicted values at time step $i$, respectively, and $n$ is the number of forecast points in a sample.

We also report the Root Mean Squared Error (RMSE), which emphasizes larger errors due to its quadratic formulation, and is defined as:

\begin{equation}
\text{RMSE} = \sqrt{ \frac{1}{n} \sum_{i=1}^{n} (y_i - \hat{y}_i)^2 }
\end{equation}

In addition, we include the Coefficient of Determination ($R^2$), which measures how well the predicted values capture the variance in the ground truth, as follows:

\begin{equation}
R^2 = 1 - \frac{ \sum_{i=1}^{n} (y_i - \hat{y}_i)^2 }{ \sum_{i=1}^{n} (y_i - \bar{y})^2 }
\end{equation}

where $\bar{y}$ is the mean of the ground truth values. An $R^2$ value close to 1 indicates that the model captures most of the variation in the data.

Although these metrics provide insight into point-wise accuracy, they do not necessarily reflect the overall quality of a forecast sequence. In applications such as hydrological forecasting, preserving the temporal shape and structure of the the time series can be more critical than minimizing absolute error. Therefore, we introduce a sequence-level metric based on standard dynamic time warping (DTW)~\cite{senin2008dynamic}, which aligns forecast and ground truth sequences before comparing them.
DTW computes the minimal cumulative distance between two time series after nonlinear alignment:

\begin{equation} \text{DTW}(X, Y) = \min_{w \in \mathcal{W}} \left( \sum_{(i,j) \in w} | x_i - y_j | \right) \end{equation}

where $X = (x_1, \ldots, x_n)$ and $Y = (y_1, \ldots, y_m)$ are the predicted and observed sequences, and $w$ is a warping path constrained by boundary and continuity conditions.
Based on the DTW-aligned sequences, we compute a sample-wise error, e.g., absolute error, which is compared against a fixed threshold $\tau$ to classify each forecast as a success or failure. The per-sample accuracy is then defined as:

\begin{equation} \text{Accuracy} = \frac{1}{N} \sum_{k=1}^{N} \mathbb{I} \left[ \text{Error}(k) < \tau \right] \end{equation}

where $N$ is the number of test samples, $\text{Error}(k)$ is the sequence error for sample $k$, and $\mathbb{I}[\cdot]$ is the indicator function.
This formulation enables a binary assessment of the quality of the forecast per test sample, capturing the overall utility of each forecast rather than just the average point deviation.
In the last step, by moving the $\tau$ over the whole range of captured error, we report the normalized area under the curve (\textit{AUC}) for the entire data set. 

\paragraph{Complexity}
We quantify each time series complexity using the \textit{statistical complexity} \(C\), grounded in the permutation entropy framework introduced by Bandt and Pompe~\cite{bandt2002permutation}, using OrdPy library~\cite{ordpy}. For each time series, we compute the ordinal pattern probability distribution \(P = \{p(\pi)\}\) of order \(d_x\). Then, we calculate the \textit{normalized permutation entropy}:
    \begin{equation}
    H = -\frac{1}{\log(d_x!)} \sum_{\pi} p(\pi)\log p(\pi), \quad 0 \le H \le 1.
    \label{eq:entropy}
    \end{equation}
It enables us to compute the \textit{Jensen--Shannon statistical complexity}:
    \begin{equation}
    C = H \cdot \frac{D_{\mathrm{JS}}(P \parallel U)}{Q_{\max}},
    \label{eq:complexity}
    \end{equation}
where \(D_{\mathrm{JS}}\) is the Jensen--Shannon divergence from the uniform distribution \(U\), and \(Q_{\max}\) is the maximum possible divergence for a given embedding dimension.

\begin{figure}[htbp]
    \centering

    \begin{subfigure}{0.75\linewidth}
        \centering
        \includegraphics[width=\linewidth]{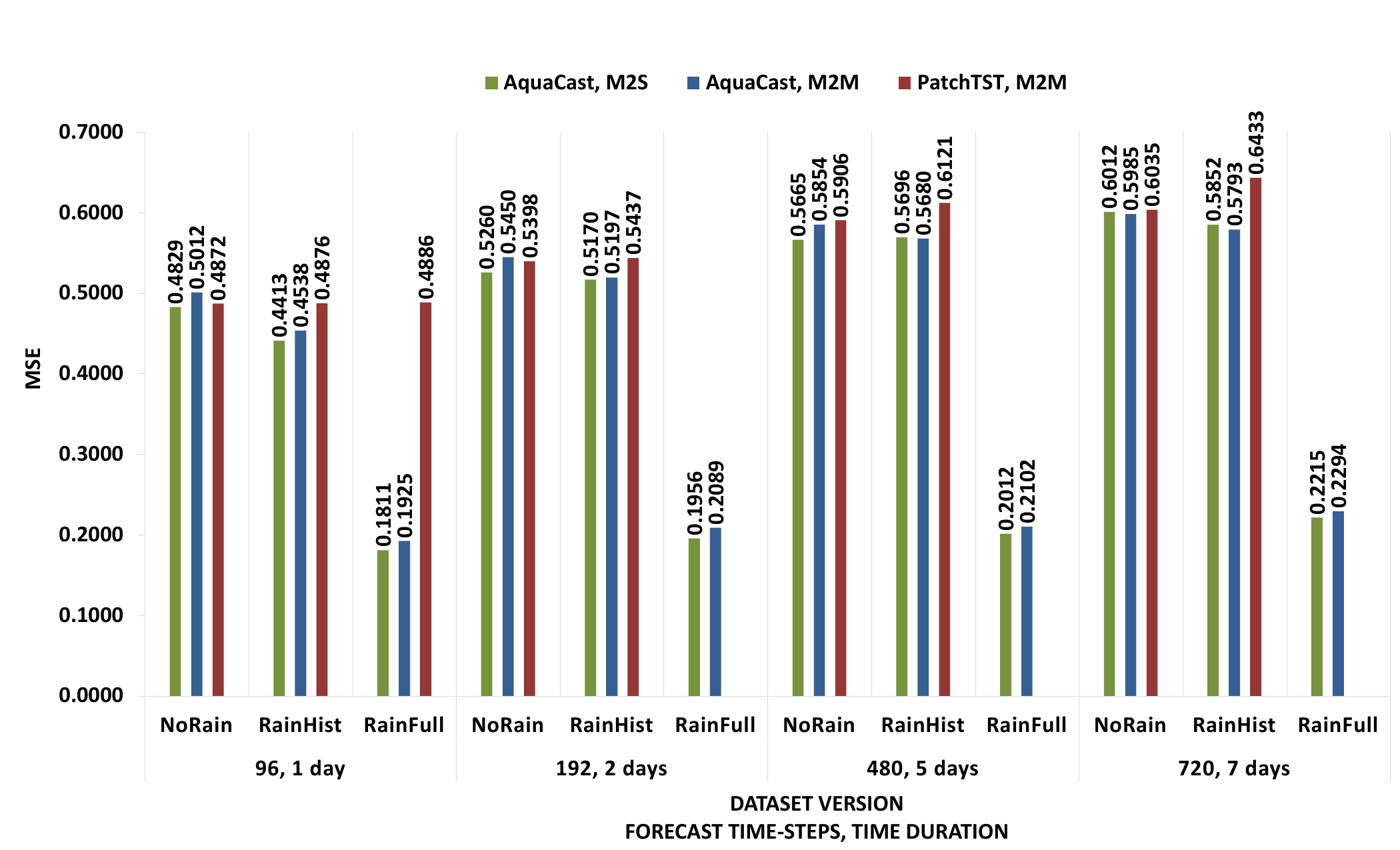}
        \label{fig:rain_info_mse}
    \end{subfigure}

    \vspace{-6.5em}

    \begin{subfigure}{0.75\linewidth}
        \centering
        \includegraphics[width=\linewidth]{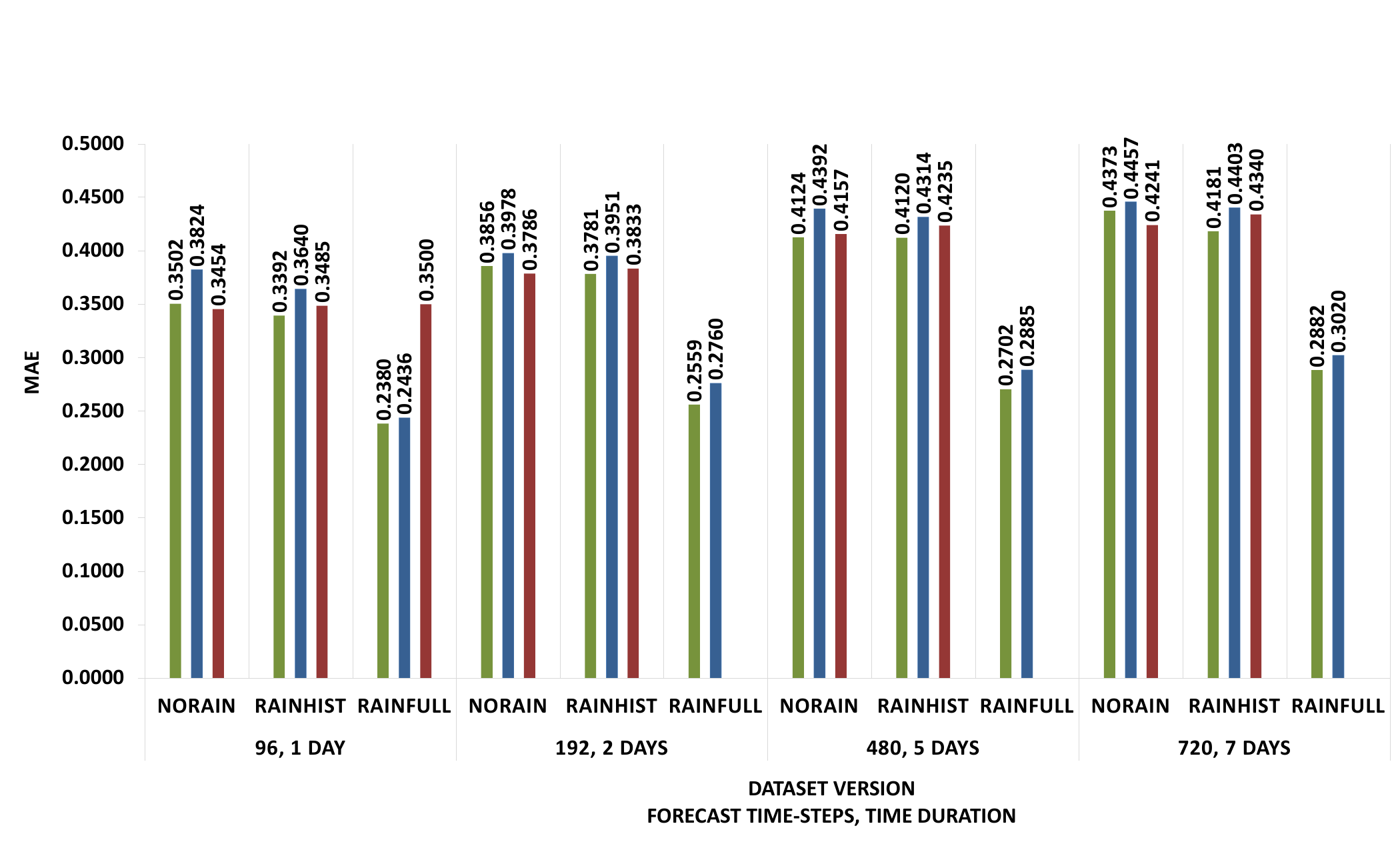}
        \label{fig:rain_info_msa}
    \end{subfigure}

    \vspace{-6.5em}

    \begin{subfigure}{0.75\linewidth}
        \centering
        \includegraphics[width=\linewidth]{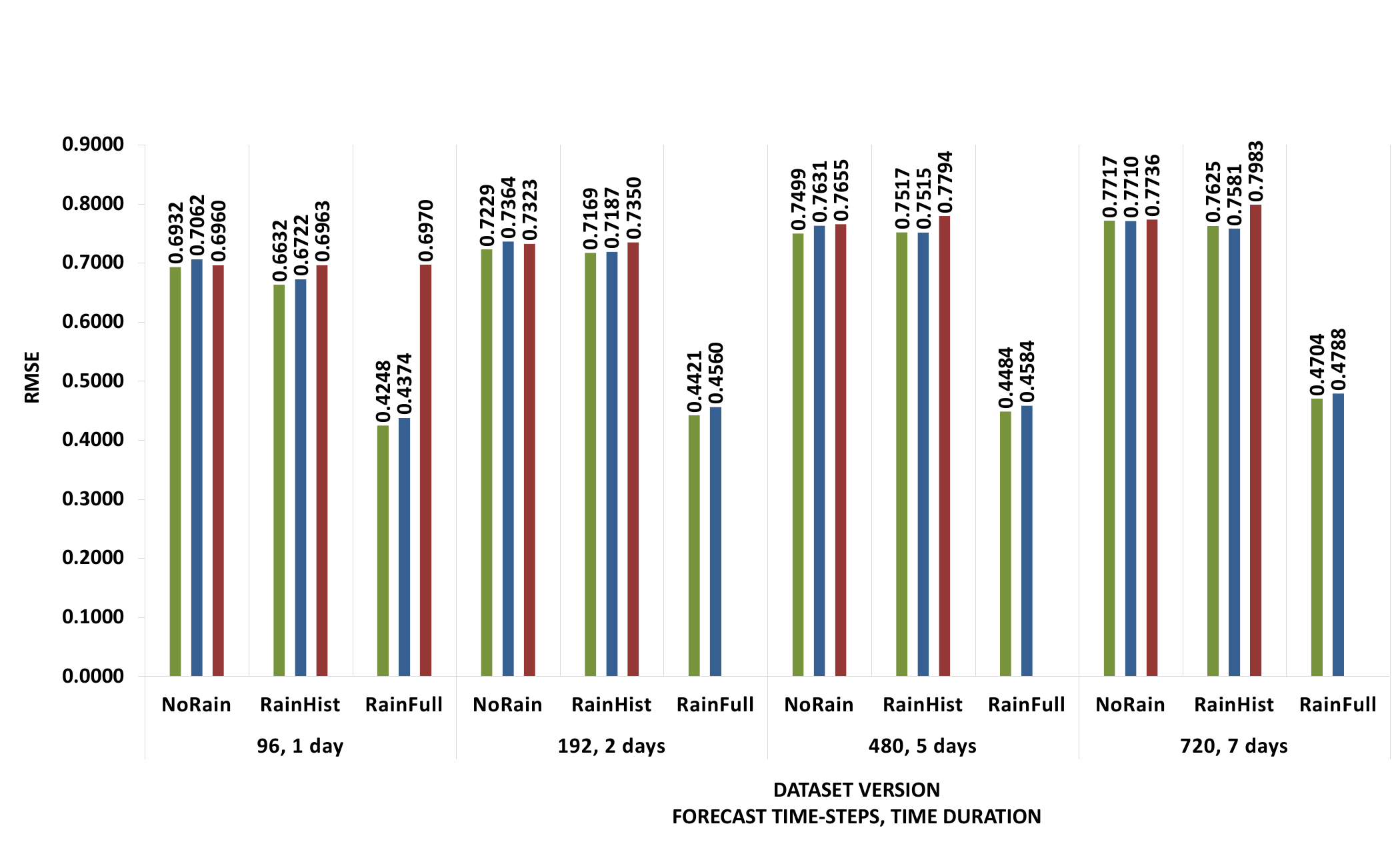}
        \label{fig:rain_info_rmse}
    \end{subfigure}
    \caption{Average error metrics of models on all 4 sensors. The history length is fixed at one day (96 time-steps). For each group of the same forecast length, the best error is achieved when both rain forecast and history (\textit{Lausanne\_RainFull}) are provided, compared to \textit{Lausanne\_NoRain} and \textit{Lausanne\_RainHist}.}
    \label{fig:rain_info_all}
\end{figure}

\begin{figure}
    \centering
    \includegraphics[width=0.6\linewidth]{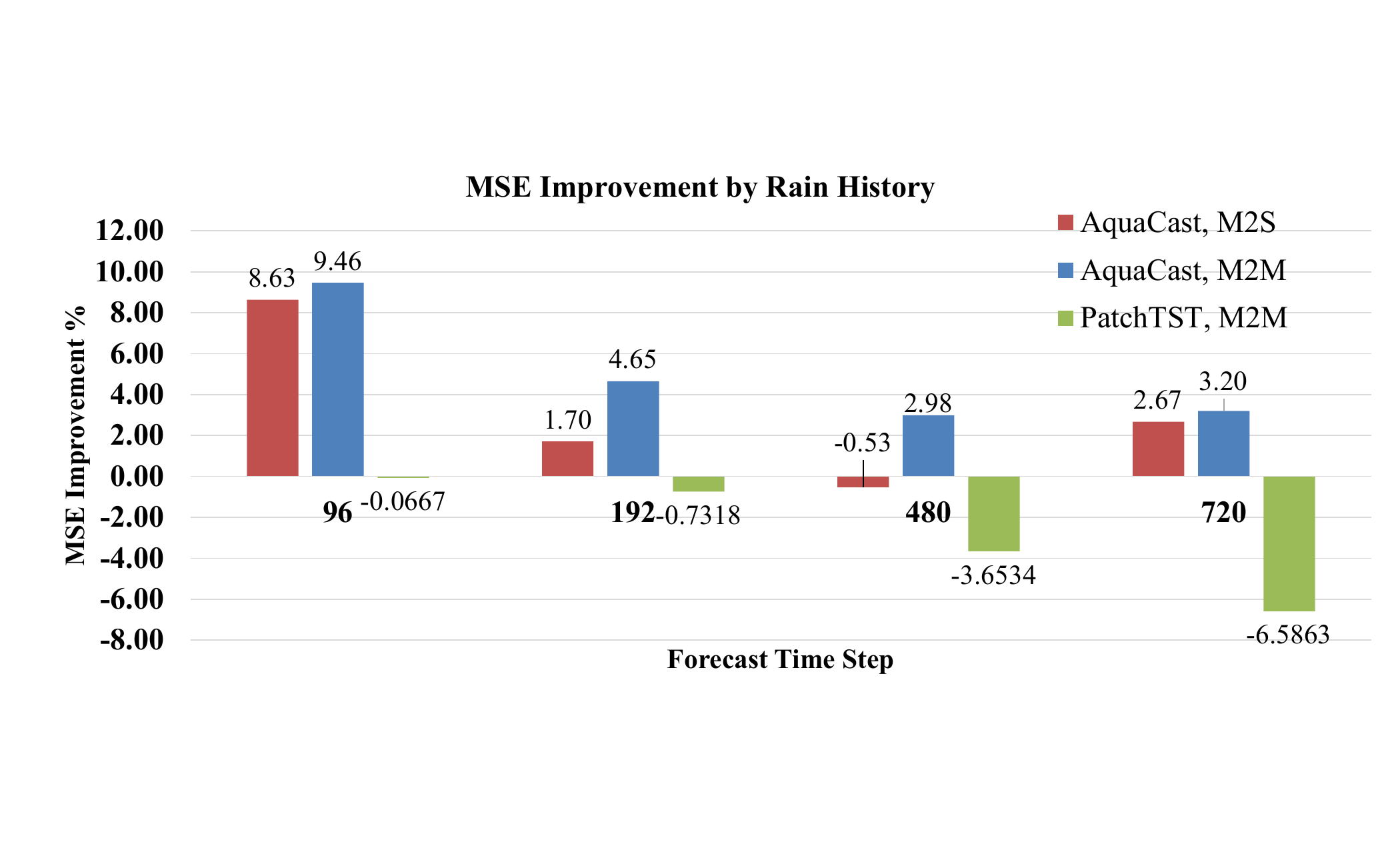}
    \vspace{-2.5em}
    \caption{Illustration of noticeable MSE improvement percentage in comparison of \textit{Lausanne\_RainHist} to \textit{Lausanne\_NoRain} for shortest forecast length. The reported percentage values are averaged over all sensors performance in LausanneCity dataset. History length is fixed for all experiments on 96 time-steps, equivalent to 1 day.}
    \label{fig:rain_history_imp}
\end{figure}

\section{Results}
\label{sec:result}

This section assesses the performance of two neural forecasting models, \MyTransformer\, in both single and multi-output configurations, compared with PatchTST~\cite{nie2022time}. It covers both real-world and synthesized multi-sensor urban water datasets. The primary aim is to understand how different rain input strategies—none, history-only, and history with forecast—affect the forecast model's ability to predict key hydrological variables. We report standard forecasting metrics, including MSE, MAE, RMSE, the coefficient of determination (R\textsuperscript{2}), and normalized AUC. The results are organized based on the real-world datasets, namely, (\textit{Lausanne\_NoRain}, \textit{Lausanne\_RainHist}, and \textit{Lausanne\_RainFull}), synthetic datasets (\textit{SynthLow}, \textit{SynthMid}, and \textit{SynthHigh}) from Table~\ref{table:dataset_version}, and visual analyses using sample forecasts to illustrate the quality of forecast.
All the experiments have a fixed history length of 96 time-steps, i.e., a full day of data at 15‑minute resolution. In this section we investigate different forecast horizons: 96, 192, 480 and 720 time-steps, which are equivalent to 1, 2, 5, and 7 days, respectively. 


\begin{figure}
    \centering
    \begin{subfigure}{0.45\linewidth}
        \centering
        \includegraphics[width=\linewidth]{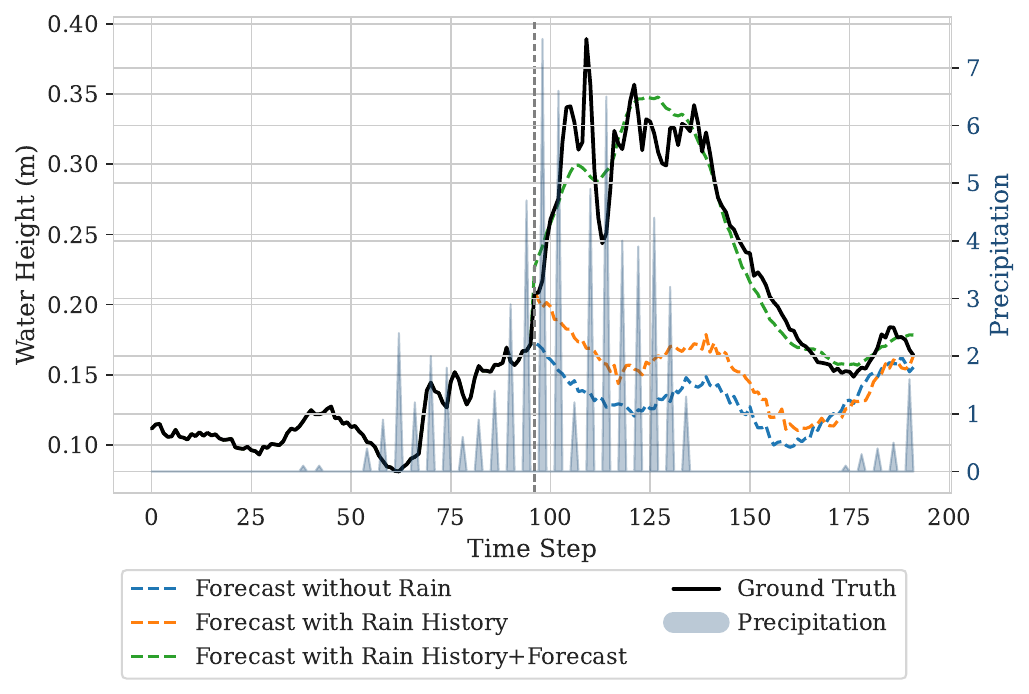}  
    \end{subfigure}
    \begin{subfigure}{0.45\linewidth}
        \centering
        \includegraphics[width=\linewidth]{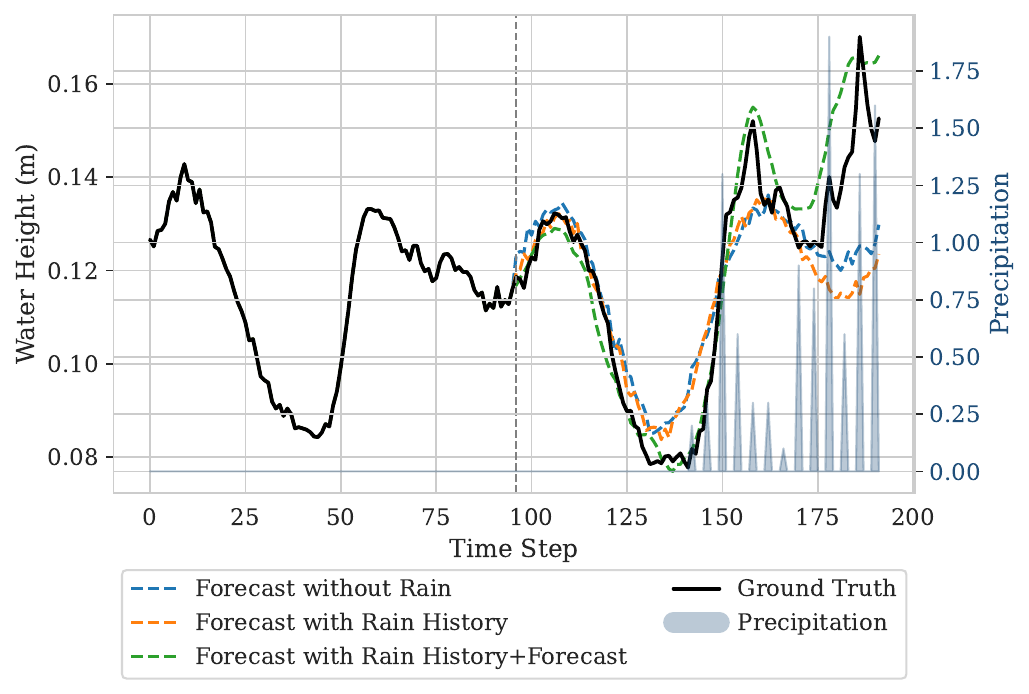}  
    \end{subfigure}
    \begin{subfigure}{0.45\linewidth}
        \centering
        \includegraphics[width=\linewidth]{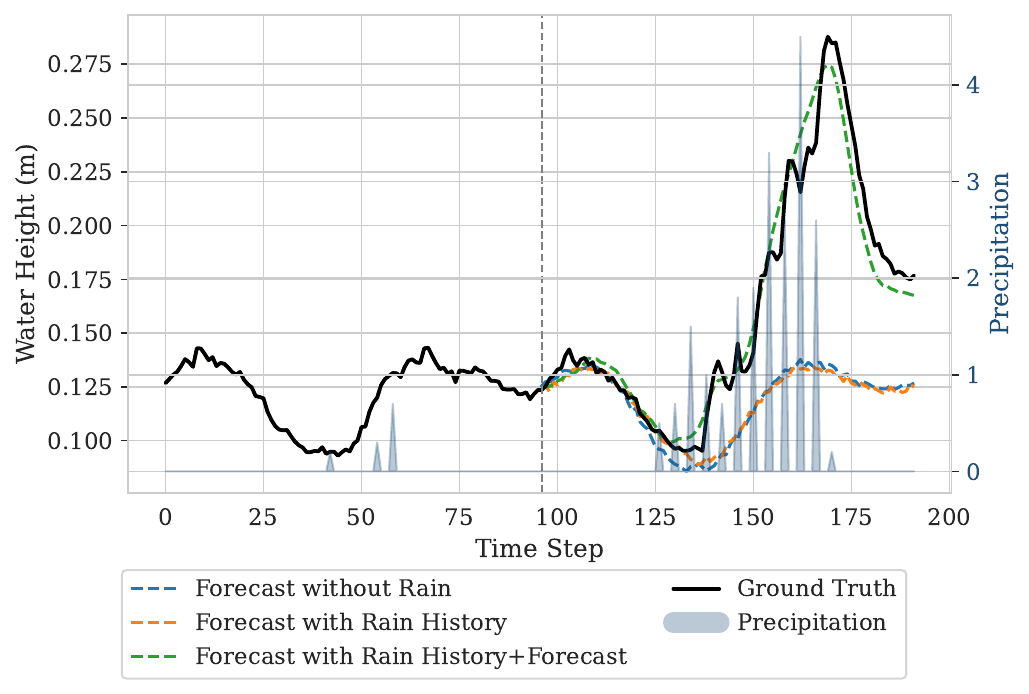}  
    \end{subfigure}
    \begin{subfigure}{0.45\linewidth}
        \centering
        \includegraphics[width=\linewidth]{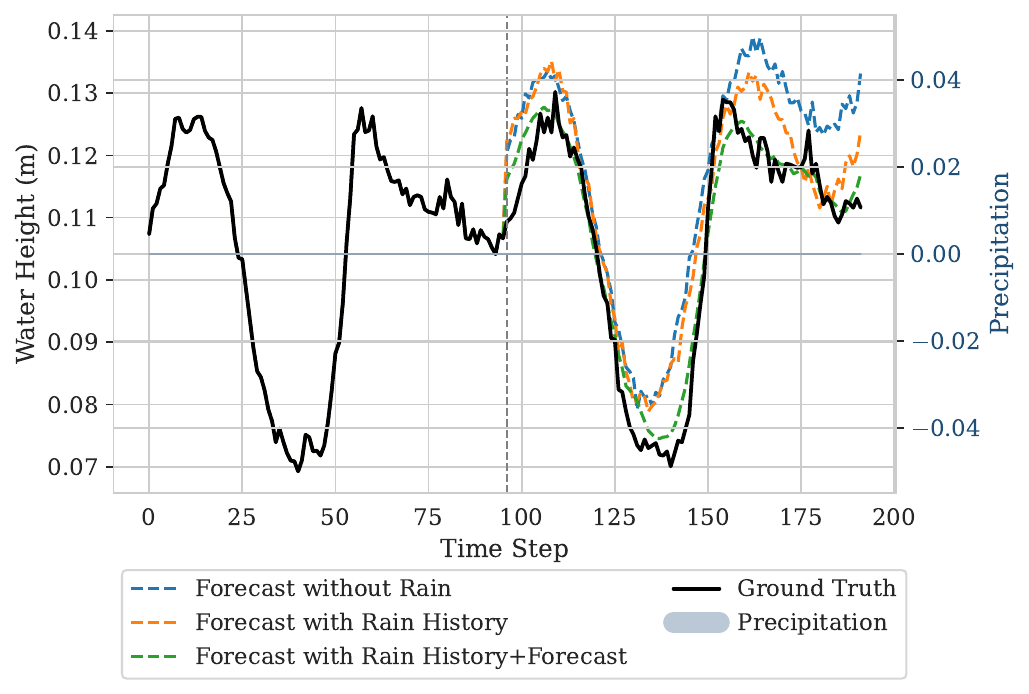}  
    \end{subfigure}
    
    \caption{\MyTransformer\ forecast samples on LausanneCity dataset.}
    \label{fig:forecast-samples-Lausanne}
\end{figure}

\begin{table}
\centering
\scriptsize
\addtolength{\tabcolsep}{-0.4em}
\begin{tabular}{lll|ccccc|ccccc|ccccc}
\hline
\multicolumn{3}{c|}{} & \multicolumn{5}{c|}{\textbf{\MyTransformer\ Multi-to-single}} & \multicolumn{5}{c|}{\textbf{\MyTransformer\ Multi-to-Multi}} & \multicolumn{5}{c}{\textbf{PatchTST Multi-to-Multi}} \\
\cline{4-18}
\textbf{\rotatebox[origin=l]{90}{Sensor}} & \textbf{\rotatebox[origin=l]{90}{Forecast}} & Data Configuration &
MSE$\downarrow$ & MAE$\downarrow$ & RMSE$\downarrow$ & R\textsuperscript{2}$\uparrow$ & AUC$\uparrow$ &
MSE$\downarrow$ & MAE$\downarrow$ & RMSE$\downarrow$ & R\textsuperscript{2}$\uparrow$ & AUC$\uparrow$ &
MSE$\downarrow$ & MAE$\downarrow$ & RMSE$\downarrow$ & R\textsuperscript{2}$\uparrow$ & AUC$\uparrow$ \\
\hline
\hline
\multirow{12}{*}{\rotatebox[origin=c]{90}{1. A}}
& \multirow{3}{*}{96}  & Lausanne\_NoRain   & 0.4901 & 0.3589 & 0.7001 & 0.3259 & 0.8573 & 0.5314 & 0.4100 & 0.7290 & 0.2701 & 0.8230 & 0.4892 & 0.3530 & 0.6994 & 0.3282 & 0.8648 \\
&                      & Lausanne\_RainHist & 0.4535 & 0.3608 & 0.6735 & 0.3779 & 0.8658 & 0.4799 & 0.3926 & 0.6928 & 0.3410 & 0.8344 & 0.4933 & 0.3604 & 0.7024 & 0.3225 & 0.8409 \\
&                      & Lausanne\_RainFull & 0.1596 & 0.2239 & 0.3995 & 0.7811 & 0.9350 & 0.1693 & 0.2343 & 0.4114 & 0.7675 & 0.9329 & 0.4947 & 0.3623 & 0.7033 & 0.3207 & 0.8656 \\
\cline{2-18}
& \multirow{3}{*}{192} & Lausanne\_NoRain   & 0.5618 & 0.4250 & 0.7495 & 0.2341 & 0.8476 & 0.5808 & 0.4297 & 0.7621 & 0.2055 & 0.8213 & 0.5659 & 0.4077 & 0.7523 & 0.2260 & 0.8203 \\
&                      & Lausanne\_RainHist & 0.5517 & 0.4104 & 0.7428 & 0.2479 & 0.8575 & 0.5623 & 0.4359 & 0.7499 & 0.2310 & 0.8445 & 0.5732 & 0.4165 & 0.7571 & 0.2161 & 0.8416 \\
&                      & Lausanne\_RainFull & 0.1920 & 0.2625 & 0.4382 & 0.7383 & 0.9266 & 0.1998 & 0.2795 & 0.4470 & 0.7267 & 0.8992 &      &        & NA       &       \\
\cline{2-18}
& \multirow{3}{*}{480} & Lausanne\_NoRain   & 0.6135 & 0.4641 & 0.7833 & 0.1768 & 0.8449 & 0.6330 & 0.4767 & 0.7956 & 0.1416 & 0.8141 & 0.6287 & 0.4583 & 0.7929 & 0.1474 & 0.8280 \\
&                      & Lausanne\_RainHist & 0.6148 & 0.4620 & 0.7841 & 0.1752 & 0.8450 & 0.6118 & 0.4720 & 0.7822 & 0.1703 & 0.8409 & 0.6531 & 0.4687 & 0.8082 & 0.1143 & 0.8240 \\
&                      & Lausanne\_RainFull & 0.1981 & 0.2700 & 0.4450 & 0.7343 & 0.9305 & 0.2126 & 0.2984 & 0.4610 & 0.7117 & 0.9221 &      &        & NA       &       \\
\cline{2-18}
& \multirow{3}{*}{720} & Lausanne\_NoRain   & 0.6460 & 0.4835 & 0.8037 & 0.1460 & 0.8387 & 0.6514 & 0.4845 & 0.8071 & 0.1239 & 0.8378 & 0.6427 & 0.4688 & 0.8017 & 0.1355 & 0.8249 \\
&                      & Lausanne\_RainHist & 0.6140 & 0.4608 & 0.7836 & 0.1884 & 0.8460 & 0.6277 & 0.4758 & 0.7923 & 0.1557 & 0.8143 & 0.6914 & 0.4765 & 0.8315 & 0.0700 & 0.8183 \\
&                      & Lausanne\_RainFull & 0.2112 & 0.2869 & 0.4596 & 0.7208 & 0.9688 & 0.2224 & 0.3066 & 0.4715 & 0.7009 & 0.9208 &      &        & NA       &       \\

\hline
\multirow{12}{*}{\rotatebox[origin=c]{90}{2. B}}
& \multirow{3}{*}{96}  & Lausanne\_NoRain   & 0.5800 & 0.3556 & 0.7616 & 0.0798 & 0.8451 & 0.5870 & 0.4001 & 0.7662 & 0.0710 & 0.8135 & 0.5929 & 0.3520 & 0.7700 & 0.0618 & 0.8648 \\
&                      & Lausanne\_RainHist & 0.5109 & 0.3282 & 0.7148 & 0.1926 & 0.8556 & 0.5305 & 0.3790 & 0.7284 & 0.1605 & 0.8236 & 0.5875 & 0.3517 & 0.7665 & 0.0703 & 0.8375 \\
&                      & Lausanne\_RainFull & 0.1757 & 0.2260 & 0.4192 & 0.7223 & 0.9378 & 0.1677 & 0.2256 & 0.4095 & 0.7347 & 0.9173 & 0.5901 & 0.3530 & 0.7682 & 0.0663 & 0.8621 \\
\cline{2-18}
& \multirow{3}{*}{192} & Lausanne\_NoRain   & 0.6124 & 0.3708 & 0.7825 & 0.0380 & 0.8415 & 0.6134 & 0.3936 & 0.7832 & 0.0330 & 0.8166 & 0.6398 & 0.3709 & 0.7999 & -0.0085 & 0.8183 \\
&                      & Lausanne\_RainHist & 0.5914 & 0.3695 & 0.7690 & 0.0711 & 0.8411 & 0.5887 & 0.3942 & 0.7673 & 0.0721 & 0.8430 & 0.6389 & 0.3718 & 0.7993 & -0.0070 & 0.8404 \\
&                      & Lausanne\_RainFull & 0.1811 & 0.2287 & 0.4255 & 0.7156 & 0.9075 & 0.1806 & 0.2412 & 0.4249 & 0.7154 & 0.8876 &      &        & NA       &        \\
\cline{2-18}
& \multirow{3}{*}{480} & Lausanne\_NoRain   & 0.6321 & 0.3894 & 0.7950 & 0.0164 & 0.8391 & 0.6252 & 0.4170 & 0.7907 & 0.0103 & 0.8151 & 0.6800 & 0.3922 & 0.8246 & -0.0764 & 0.8213 \\
&                      & Lausanne\_RainHist & 0.6385 & 0.3895 & 0.7991 & 0.0065 & 0.8379 & 0.6122 & 0.4185 & 0.7824 & 0.0310 & 0.8419 & 0.7006 & 0.4007 & 0.8370 & -0.1089 & 0.8178 \\
&                      & Lausanne\_RainFull & 0.1888 & 0.2459 & 0.4345 & 0.7063 & 0.9197 & 0.2087 & 0.2670 & 0.4568 & 0.6697 & 0.9120 &      &        & NA       &        \\
\cline{2-18}
& \multirow{3}{*}{720} & Lausanne\_NoRain   & 0.6631 & 0.4194 & 0.8143 & -0.0156 & 0.8301 & 0.6428 & 0.4281 & 0.8018 & -0.0049 & 0.8340 & 0.6945 & 0.3962 & 0.8334 & -0.0857 &0.8164 \\
&                      & Lausanne\_RainHist & 0.6456 & 0.3946 & 0.8035 & 0.0113  & 0.8395 & 0.6377 & 0.4321 & 0.7986 & 0.0031  & 0.8113 & 0.7321 & 0.4090 & 0.8556 & -0.1444 & 0.8085 \\
&                      & Lausanne\_RainFull & 0.2044 & 0.2549 & 0.4521 & 0.6869  & 0.9123 & 0.2219 & 0.2746 & 0.4711 & 0.6530  & 0.9097 &      &        & NA       &        \\
\hline
\multirow{12}{*}{\rotatebox[origin=c]{90}{3. C}}
& \multirow{3}{*}{96}  & Lausanne\_NoRain   & 0.5800 & 0.3556 & 0.7616 & 0.0798 & 0.8451 & 0.5870 & 0.4001 & 0.7662 & 0.0710 & 0.8135 & 0.5929 & 0.3520 & 0.7700 & 0.0618 & 0.8440 \\
&                      & Lausanne\_RainHist & 0.5109 & 0.3282 & 0.7148 & 0.1926 & 0.8556 & 0.5305 & 0.3790 & 0.7284 & 0.1605 & 0.8236 & 0.5875 & 0.3517 & 0.7665 & 0.0703 & 0.8199 \\
&                      & Lausanne\_RainFull & 0.1757 & 0.2260 & 0.4192 & 0.7223 & 0.9378 & 0.1677 & 0.2256 & 0.4095 & 0.7347 & 0.9173 & 0.5901 & 0.3530 & 0.7682 & 0.0663 & 0.8445 \\
\cline{2-18}
& \multirow{3}{*}{192} & Lausanne\_NoRain   & 0.6124 & 0.3708 & 0.7825 & 0.0380 & 0.8415 & 0.6134 & 0.3936 & 0.7832 & 0.0330 & 0.8166 & 0.6398 & 0.3709 & 0.7999 & -0.0085 & 0.8107 \\
&                      & Lausanne\_RainHist & 0.5914 & 0.3695 & 0.7690 & 0.0711 & 0.8411 & 0.5887 & 0.3942 & 0.7673 & 0.0721 & 0.8430 & 0.6389 & 0.3718 & 0.7993 & -0.0070 & 0.8352 \\
&                      & Lausanne\_RainFull & 0.1811 & 0.2287 & 0.4255 & 0.7156 & 0.9075 & 0.1806 & 0.2412 & 0.4249 & 0.7154 & 0.8876 &      &        & NA       &        \\
\cline{2-18}
& \multirow{3}{*}{480} & Lausanne\_NoRain   & 0.6321 & 0.3894 & 0.7950 & 0.0164 & 0.8391 & 0.6252 & 0.4170 & 0.7907 & 0.0103 & 0.8151 & 0.6800 & 0.3922 & 0.8246 & -0.0764 & 0.8308 \\
&                      & Lausanne\_RainHist & 0.6385 & 0.3895 & 0.7991 & 0.0065 & 0.8379 & 0.6122 & 0.4185 & 0.7824 & 0.0310 & 0.8419 & 0.7006 & 0.4007 & 0.8370 & -0.1089 & 0.8267 \\
&                      & Lausanne\_RainFull & 0.1888 & 0.2459 & 0.4345 & 0.7063 & 0.9197 & 0.2087 & 0.2670 & 0.4568 & 0.6697 & 0.9120 &      &        & NA       &        \\
\cline{2-18}
& \multirow{3}{*}{720} & Lausanne\_NoRain   & 0.6631 & 0.4194 & 0.8143 & -0.0156 & 0.8301 & 0.6428 & 0.4281 & 0.8018 & -0.0049 & 0.8377 & 0.6945 & 0.3962 & 0.8334 & -0.0857 & 0.8273 \\
&                      & Lausanne\_RainHist & 0.6456 & 0.3946 & 0.8035 & 0.0113  & 0.8395 & 0.6377 & 0.4321 & 0.7986 & 0.0031  & 0.8113 & 0.7321 & 0.4090 & 0.8556 & -0.1444 & 0.8205 \\
&                      & Lausanne\_RainFull & 0.2044 & 0.2549 & 0.4521 & 0.6869  & 0.9123 & 0.2219 & 0.2746 & 0.4711 & 0.6530  & 0.9097 &      &        & NA       &        \\
\hline
\multirow{12}{*}{\rotatebox[origin=c]{90}{4. D}}
& \multirow{3}{*}{96}  & Lausanne\_NoRain   & 0.3878 & 0.3354 & 0.6228 & 0.0767 & 0.8521 & 0.3947 & 0.3414 & 0.6283 & 0.0635 & 0.8280 & 0.3832 & 0.3157 & 0.6191 & 0.0909 & 0.8478 \\
&                      & Lausanne\_RainHist & 0.3696 & 0.3197 & 0.6079 & 0.1217 & 0.8614 & 0.3704 & 0.3363 & 0.6086 & 0.1215 & 0.8342 & 0.3818 & 0.3166 & 0.6179 & 0.0945 & 0.8230 \\
&                      & Lausanne\_RainFull & 0.2184 & 0.2646 & 0.4673 & 0.4810 & 0.9326 & 0.2461 & 0.2704 & 0.4961 & 0.4163 & 0.8893 & 0.3818 & 0.3173 & 0.6179 & 0.0944 & 0.8484 \\
\cline{2-18}
& \multirow{3}{*}{192} & Lausanne\_NoRain   & 0.3936 & 0.3352 & 0.6273 & 0.0701 & 0.8582 & 0.4184 & 0.3514 & 0.6469 & 0.0142 & 0.8311 & 0.4042 & 0.3271 & 0.6358 & 0.0479 & 0.8227 \\
&                      & Lausanne\_RainHist & 0.3905 & 0.3301 & 0.6249 & 0.0775 & 0.8566 & 0.3897 & 0.3365 & 0.6243 & 0.0821 & 0.8610 & 0.4060 & 0.3286 & 0.6372 & 0.0438 & 0.8465 \\
&                      & Lausanne\_RainFull & 0.2088 & 0.2650 & 0.4569 & 0.5069 & 0.9329 & 0.2554 & 0.3089 & 0.5053 & 0.3986 & 0.8793 &      &        & NA       &       \\
\cline{2-18}
& \multirow{3}{*}{480} & Lausanne\_NoRain   & 0.4069 & 0.3452 & 0.6379 & 0.0477 & 0.8594 & 0.4448 & 0.3936 & 0.6669 & -0.0483 & 0.8303 & 0.4227 & 0.3412 & 0.6501 & 0.0038 & 0.8462 \\
&                      & Lausanne\_RainHist & 0.4033 & 0.3437 & 0.6350 & 0.0564 & 0.8553 & 0.4279 & 0.3722 & 0.6541 & -0.0085 & 0.8580 & 0.4377 & 0.3458 & 0.6616 & -0.0318 & 0.8426 \\
&                      & Lausanne\_RainFull & 0.1990 & 0.2712 & 0.4461 & 0.5344 & 0.9339 & 0.1952 & 0.2792 & 0.4419 & 0.5398 & 0.8916 &      &        & NA       &       \\
\cline{2-18}
& \multirow{3}{*}{720} & Lausanne\_NoRain   & 0.4127 & 0.3536 & 0.6424 & 0.0488 & 0.8574 & 0.4361 & 0.3939 & 0.6604 & -0.0304 & 0.8578 & 0.4243 & 0.3431 & 0.6514 & -0.0024 & 0.8443 \\
&                      & Lausanne\_RainHist & 0.4308 & 0.3483 & 0.6564 & 0.0073 & 0.8593 & 0.4113 & 0.3811 & 0.6413 & 0.0284 & 0.8280 & 0.4418 & 0.3486 & 0.6647 & -0.0438 & 0.8399 \\
&                      & Lausanne\_RainFull & 0.2250 & 0.2909 & 0.4743 & 0.4816 & 0.9329 & 0.2245 & 0.2977 & 0.4738 & 0.4696 & 0.8856 &      &        & NA       &       \\
\hline
\hline
\multirow{12}{*}{\rotatebox[origin=c]{90}{Mean}}
 & \multirow{3}{*}{96} & Lausanne\_NoRain   & 0.4829 & 0.3502 & 0.6932 & 0.2133 &       & 0.5012 & 0.3824 & 0.7062 & 0.1863 &       & 0.4872 & 0.3454 & 0.6960 & 0.2081 & 0.8566 \\
&& Lausanne\_RainHist & 0.4413 & 0.3392 & 0.6632 & 0.2786 &       & 0.4538 & 0.3640 & 0.6722 & 0.2602 &       & 0.4876 & 0.3485 & 0.6963 & 0.2084 & 0.8631 \\
&& Lausanne\_RainFull & 0.1811 & 0.2380 & 0.4248 & 0.6889 &       & 0.1925 & 0.2436 & 0.4374 & 0.6669 &       & 0.4886 & 0.3500 & 0.6970 & 0.2068 & 0.8681 \\
 \cline{2-18}
 & \multirow{3}{*}{192} & Lausanne\_NoRain   & 0.5260 & 0.3856 & 0.7229 & 0.1566 &       & 0.5450 & 0.3978 & 0.7364 & 0.1232 &       & 0.5398 & 0.3786 & 0.7323 & 0.1325 & 0.8433 \\
&& Lausanne\_RainHist & 0.5170 & 0.3781 & 0.7169 & 0.1707 &       & 0.5197 & 0.3951 & 0.7187 & 0.1662 &       & 0.5437 & 0.3833 & 0.7350 & 0.1268 & 0.8534 \\
&& Lausanne\_RainFull & 0.1956 & 0.2559 & 0.4421 & 0.6733 &       & 0.2089 & 0.2760 & 0.4560 & 0.6434 &       &        &        & NA       &       \\
 \cline{2-18}
 & \multirow{3}{*}{480} & Lausanne\_NoRain   & 0.5665 & 0.4124 & 0.7499 & 0.1072 &       & 0.5854 & 0.4392 & 0.7631 & 0.0627 &       & 0.5906 & 0.4157 & 0.7655 & 0.0581 & 0.8316 \\
&& Lausanne\_RainHist & 0.5696 & 0.4120 & 0.7517 & 0.1039 &       & 0.5680 & 0.4314 & 0.7515 & 0.0913 &       & 0.6121 & 0.4235 & 0.7794 & 0.0241 & 0.8444 \\
&& Lausanne\_RainFull & 0.2012 & 0.2702 & 0.4484 & 0.6713 &       & 0.2102 & 0.2885 & 0.4584 & 0.6554 &       &        &        & NA       &       \\
 \cline{2-18}
 & \multirow{3}{*}{720} & Lausanne\_NoRain   & 0.6012 & 0.4373 & 0.7717 & 0.0698 &       & 0.5985 & 0.4457 & 0.7710 & 0.0512 &       & 0.6035 & 0.4241 & 0.7736 & 0.0445 & 0.8282 \\
&& Lausanne\_RainHist & 0.5852 & 0.4181 & 0.7625 & 0.0876 &       & 0.5793 & 0.4403 & 0.7581 & 0.0834 &       & 0.6433 & 0.4340 & 0.7983 & -0.0153 & 0.8426 \\
&& Lausanne\_RainFull & 0.2215 & 0.2882 & 0.4704 & 0.6415 &       & 0.2294 & 0.3020 & 0.4788 & 0.6231 &       &        &        & NA       &       \\
\hline
\end{tabular}
\caption{Comparison of \MyTransformer\ and PatchTST across Sensors, Dataset Versions, and Forecast Lengths on LausanneCity dataset. PatchTST is not applicable (NA) to train on data configuration \textit{Lausanne\_RainFull} which has future precipitation values where forecast length is not equal to history length. History length is 96 time-steps for all results.}
\label{result:myTransformerAll}
\end{table}

\subsection{Performance Evaluation on the Lausanne City Dataset}
\label{subsec:real}
The LausanneCity datasets, summarized in Table~\ref{table:dataset_version}, offer increasing information sharing in rain-related input time-series: \textit{Lausanne\_NoRain} includes only endogenous time-series from Table~\ref{table:sensor_overview}; utilizing MeteoSwiss~\cite{meteoswiss} precipitation records; \textit{Lausanne\_RainHist} adds historical rainfall and \textit{Lausanne\_RainFull} incorporates both rainfall history and perfect forecast from the same MeteoSwiss records. These design choices reflect the practical observation that precipitation impacts on urban water dynamics occur with minimal lag, specifically in steep urban areas such as Lausanne city, making the forecast reports potentially valuable.

Table~\ref{result:myTransformerAll} presents the detailed performance of PatchTST~\cite{nie2022time} and \MyTransformer\ in different versions of the LausanneCity dataset on all sensors. PatchTST performs best on \textit{Lausanne\_NoRain} configuration, with a noticeable decline as additional rainfall information is incorporated. For example, error values across all sensors remain relatively unchanged from \textit{Lausanne\_NoRain} to \textit{Lausanne\_RainHist} and \textit{Lausanne\_RainFull}, despite the rain information sharing. This drop in performance can be attributed to PatchTST’s channel-independent architecture, which processes each time series separately using shared weights, limiting its ability to capture cross-channel dependencies such as the influence of rainfall on downstream water dynamics. Moreover, due to its multi-to-multi design, the model learns to predict both target and exogenous variables when additional inputs like rain history and forecasts are provided, potentially diluting its focus on water-related targets. Furthermore, PatchTST lacks support for variable history and forecast lengths, which restricts its effectiveness in settings like \textit{Lausanne\_RainFull}, where precipitation forecasts introduce this variability.

\MyTransformer\ is presented in multi-to-single and multi-to-multi configurations in Table~\ref{result:myTransformerAll}. Conversely, the results show strong and consistent improvements across dataset versions for all sensors. Also on average of all sensors, Figure~\ref{fig:rain_info_all} illustrates that the model effectively utilizes rain inputs, particularly in \textit{Lausanne\_RainFull}, where both history and forecast are accessible. For instance, sensor 1.~\textit{A} reached the lowest error values, among all sensors for all forecast length in single output configuration with 96 time-steps forecast length. In single output configuration, by using the full rain information, sensor 3.~\textit{C} reached the highest MSE improvement from 0.6124 to 0.1811 (70.42\%) for 192 forecast length and sensor 4.~\textit{D} reached the minimum MSE improvement from 0.3878 to 0.2184 (43.68\%) for 96 time-steps, corresponding to 1 day forecast. Switching to multi-output configuration to predict all sensors with one single model utilizing the \textit{Lausanne\_RainFull}, has not changed the results significantly - on average of sensors, MSE degrade between 6.77\% (from 0.1956 to 0.2089) to 3.54\% (from 0.2215 to 0.2294) among all forecast lengths. The same trend is also captured using other error metrics, MAE and RMSE. Normalized AUC values, from Table~\ref{result:myTransformerAll}, also rise with increasing shared rain information, e.g. sensor 1.~\textit{A} that normalized AUC improve from 0.8230 to 0.9329. These findings demonstrate enhanced shape alignment in the model’s forecasts, reflected in lower errors and an increased accuracy AUC. These improvements reflect the model's ability to learn inter-variable dependencies through its attention-based architecture.


Another notable outcome is the effect of forecast horizon and precipitation information on model performance. As shown in Figure~\ref{fig:rain_history_imp}, when the forecast length is short (e.g., 96), incorporating only rain history leads to the most significant improvement. For example, using the \textit{Lausanne\_RainHist} dataset instead of \textit{Lausanne\_NoRain}, \MyTransformer\ in the multi-to-multi setup achieves a 9.46\% reduction in MSE, from 0.5012 to 0.4538. In contrast, for a longer forecast horizon of 720 time steps under the same conditions, MSE reduces only by 3.2\%, from 0.5985 to 0.5793.
This suggests that \MyTransformer, which is designed to capture peak event sources effectively, benefits more from rain history when the target peak lies closer in time. In such cases, the relevant precipitation time-series is already present in the historical data and can be immediately reflected in water feature dynamics.

Figure~\ref{fig:forecast-samples-Lausanne} provides sample forecast plots that illustrate how \MyTransformer\ captures temporal dynamics under the three rainfall configurations. In the \textit{Lausanne\_NoRain}, forecasts generally follow trends, but exhibit timing errors near rapid changes. In contrast, the history-only setup (\textit{Lausanne\_RainHist}) slightly improves reactivity to rainfall-induced spikes. The best alignment is seen in \textit{Lausanne\_RainFull}, where both the timing and the amplitude of peak responses are well captured. This visual evidence complements the quantitative metrics and highlights the importance of integrating forecast information in real-time applications.

These findings highlight why \MyTransformer\ consistently outperforms PatchTST. Its attention-based architecture is specifically designed to capture both temporal patterns and inter-variable relationships, enabling it to model the causal effects of rainfall on downstream water dynamics, an essential feature in urban hydrological systems. Unlike PatchTST, which treats each time series independently, \MyTransformer\ jointly processes multi-source inputs and effectively learns contextual dependencies between exogenous and endogenous variables. This design allows it to adaptively focus on the most relevant time series, such as upcoming rainfall surges, leading to more precise and timely forecasts. The flexibility of the model in handling varying input and output lengths also enhances its robustness when real-world constraints demand forecasting under different time horizons and data configurations.

\begin{figure}[htbp]
    \centering

    \begin{subfigure}[b]{0.32\textwidth}
        \centering
        \includegraphics[width=\textwidth]{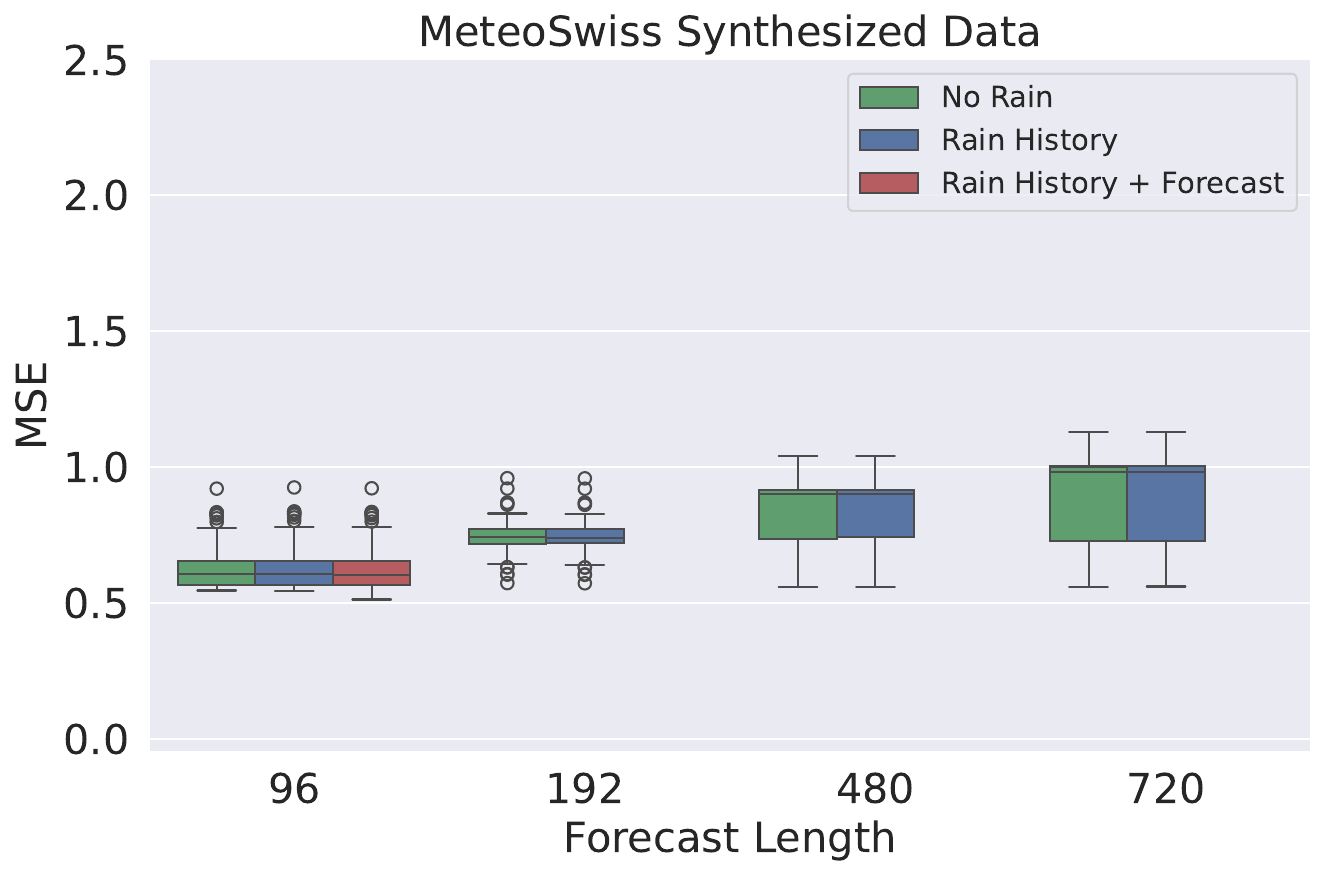}
        \caption{PatchTST on \textit{SynthLow}}
        \label{fig:box-synthL-patchtst}
    \end{subfigure}
    \hfill
    \begin{subfigure}[b]{0.32\textwidth}
        \centering
        \includegraphics[width=\textwidth]{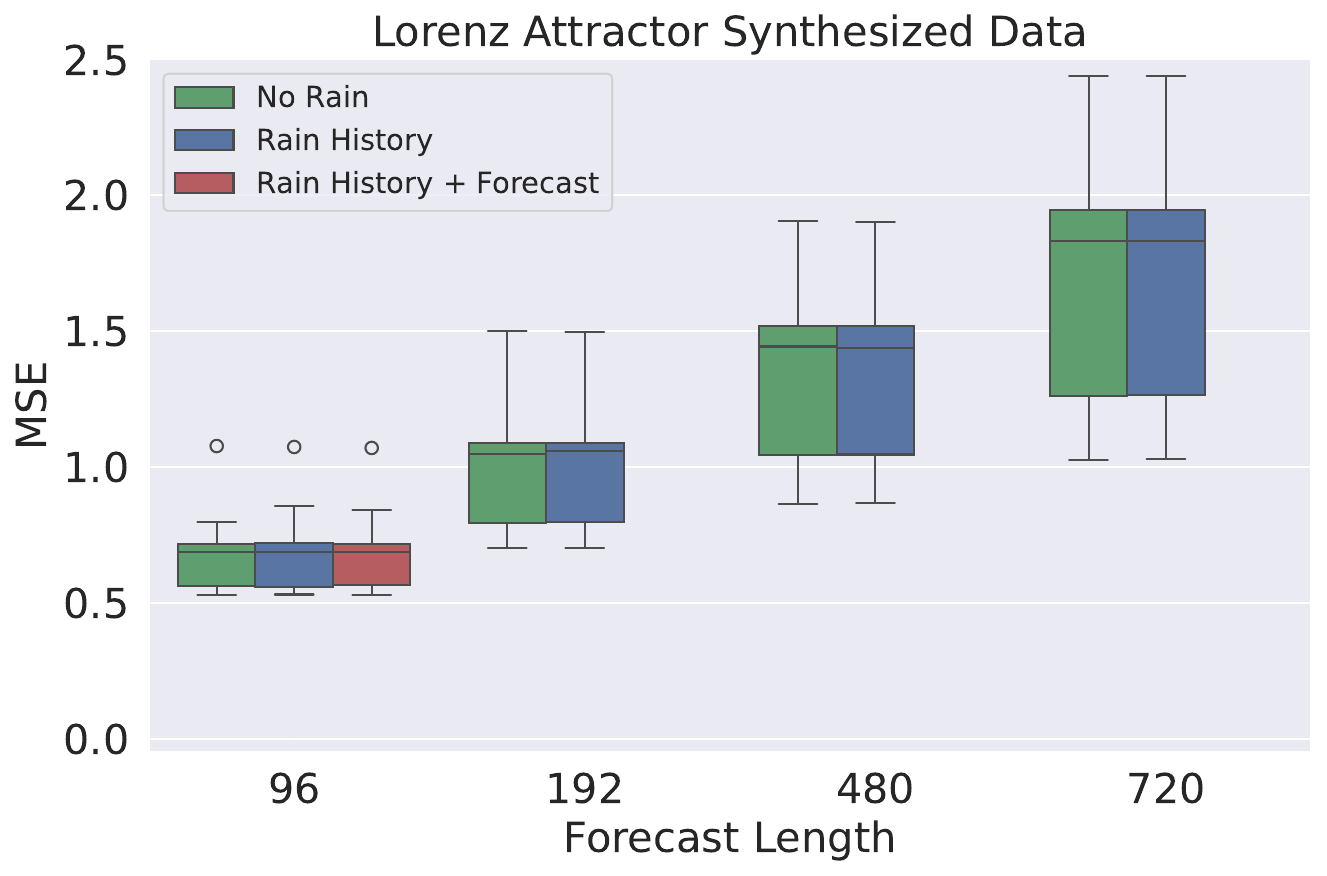}
        \caption{PatchTST on \textit{SynthMid}}
        \label{fig:box-synthM-patchtst}
    \end{subfigure}
    \hfill
    \begin{subfigure}[b]{0.32\textwidth}
        \centering
        \includegraphics[width=\textwidth]{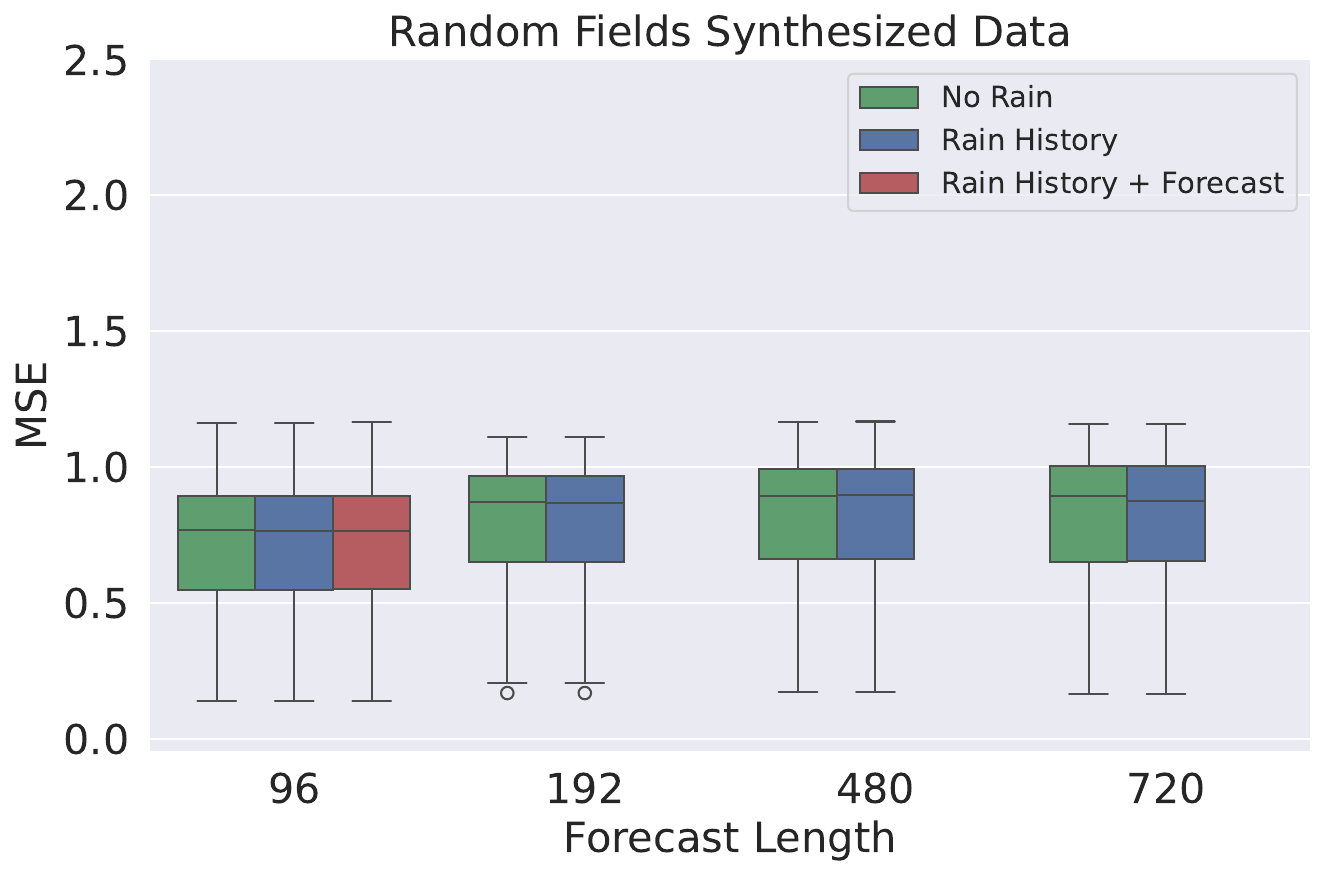}
        \caption{PatchTST on \textit{SynthHigh}}
        \label{fig:box-synthH-patchtst}
    \end{subfigure}
    
    \vspace{1em}
    

    \begin{subfigure}[b]{0.32\textwidth}
        \centering
        \includegraphics[width=\textwidth]{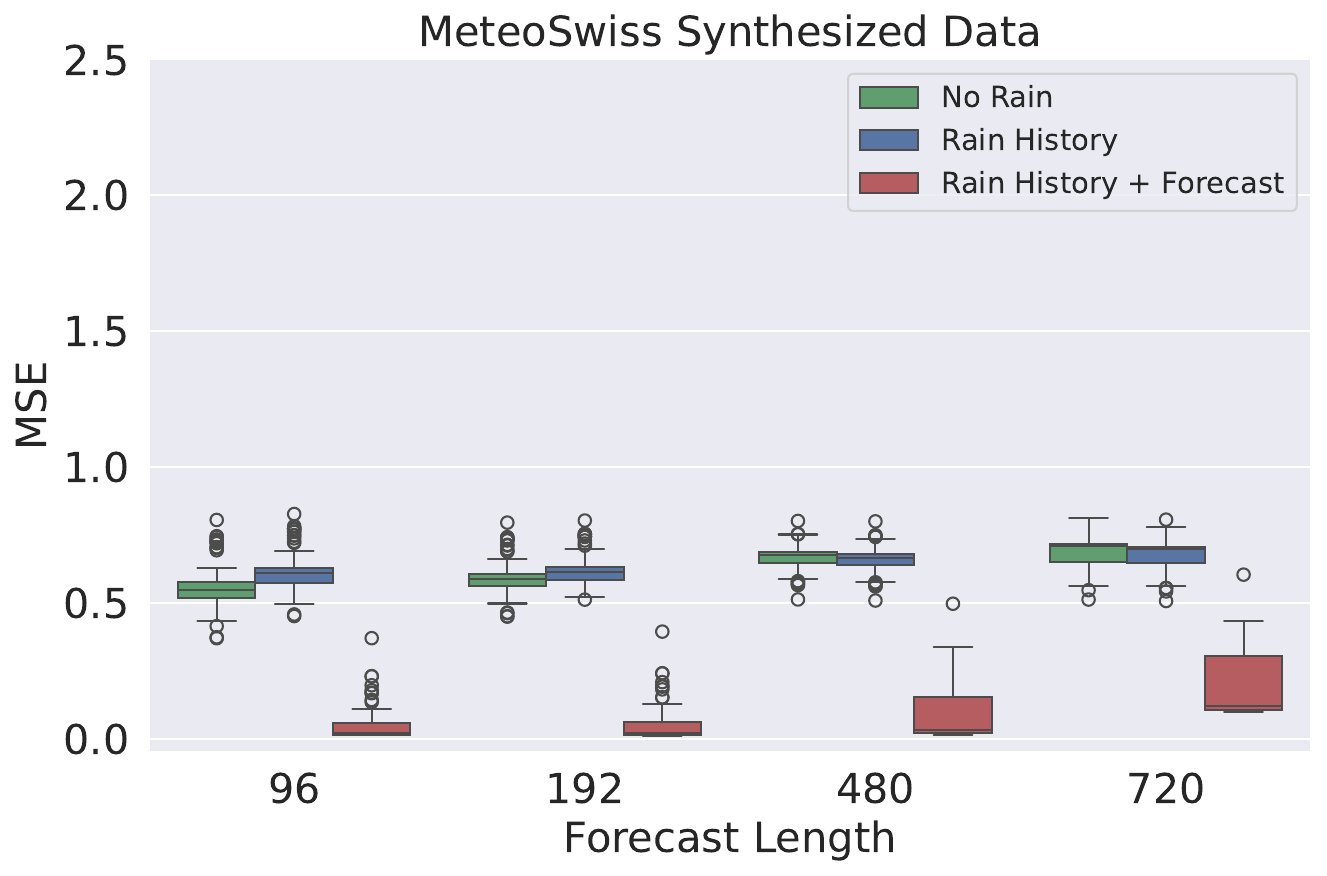}
        \caption{\MyTransformer\ on \textit{SynthLow}}
        \label{fig:box-synthL-myT}
        \end{subfigure}
    \hfill
    \begin{subfigure}[b]{0.32\textwidth}
        \centering
        \includegraphics[width=\textwidth]{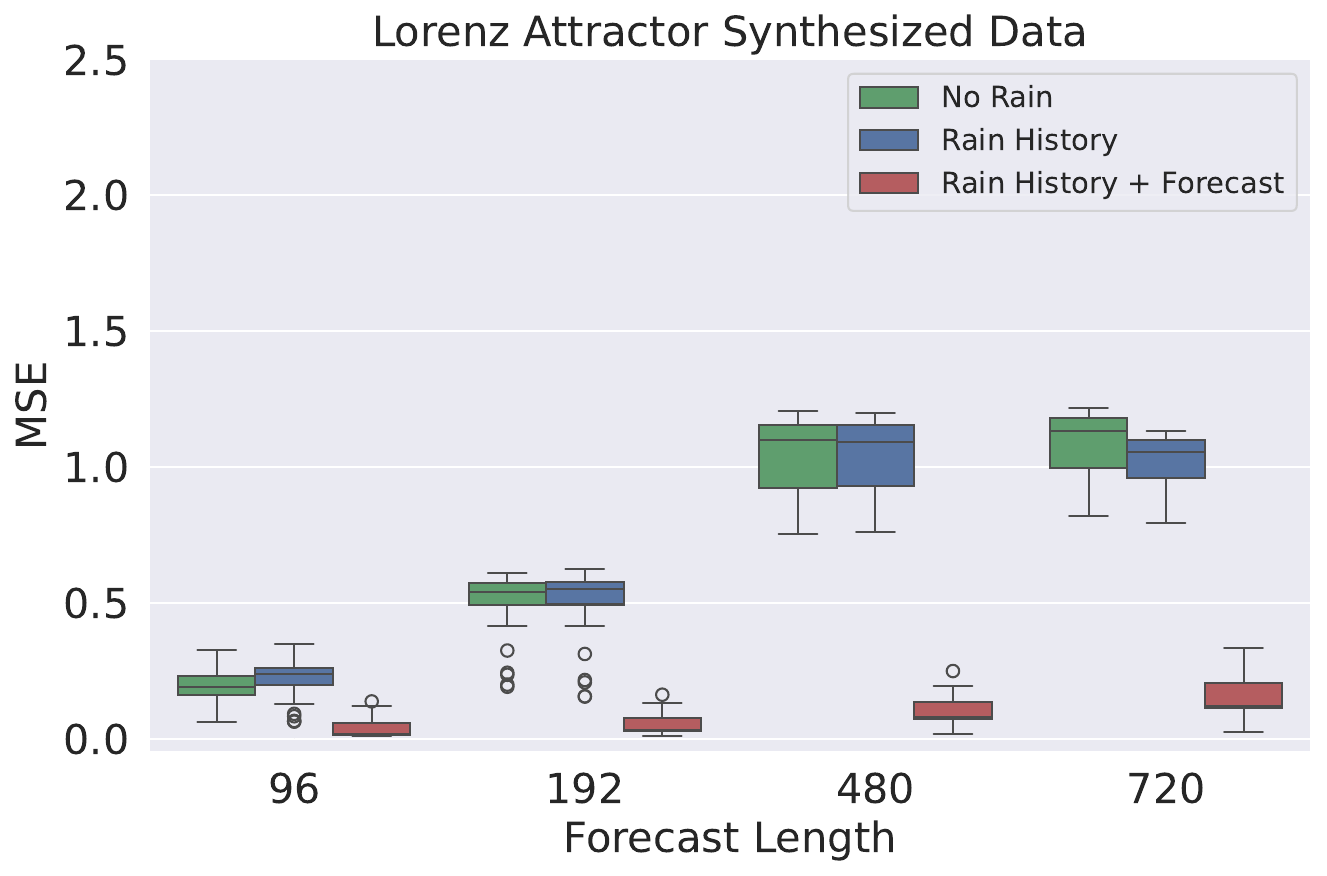}
        \caption{\MyTransformer\ on \textit{SynthMid}}
        \label{fig:box-synthM-myT}
    \end{subfigure}
    \hfill
    \begin{subfigure}[b]{0.32\textwidth}
        \centering
        \includegraphics[width=\textwidth]{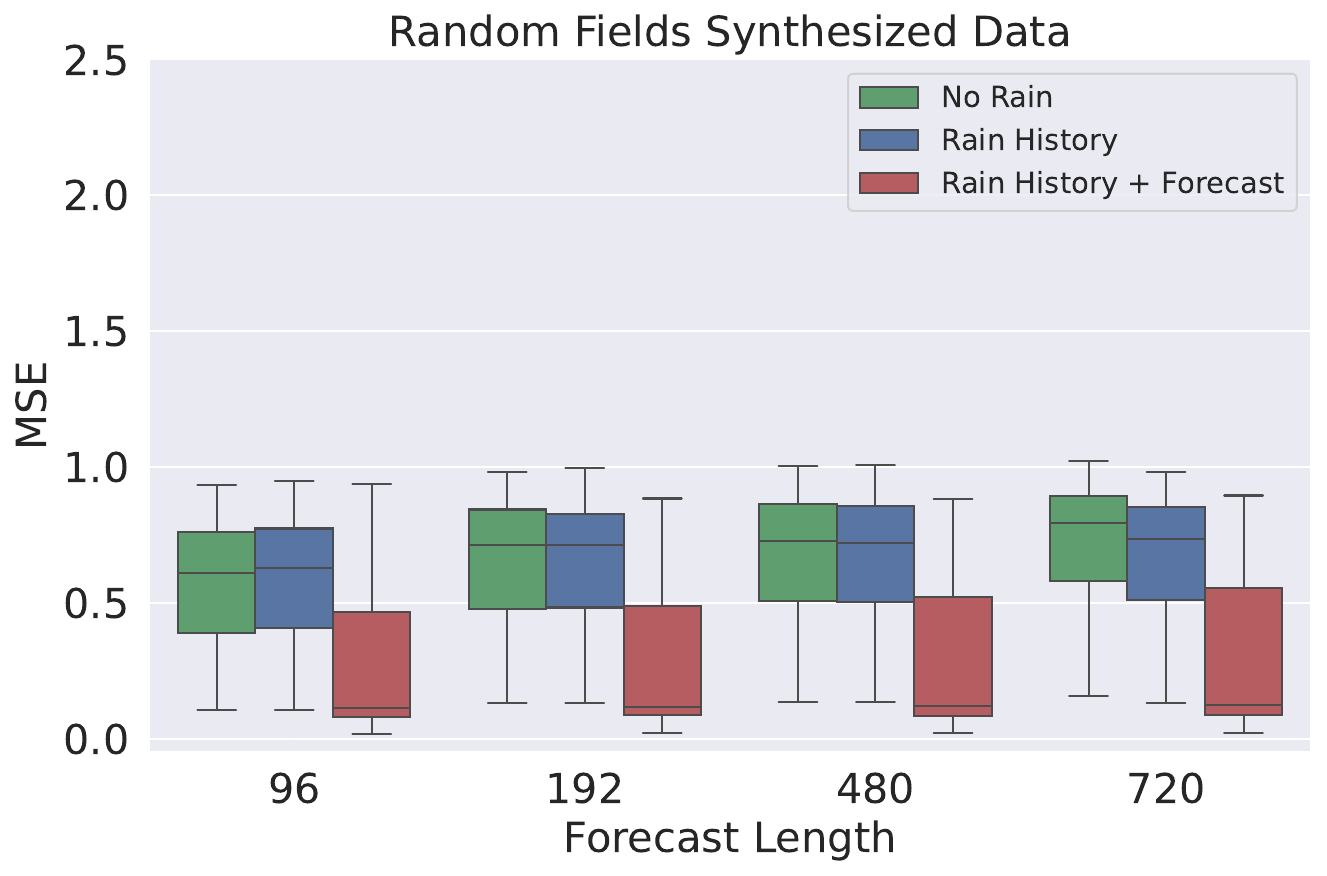}
        \caption{\MyTransformer\ on \textit{SynthHigh}}
        \label{fig:box-synthH-myT}
    \end{subfigure}

    \caption{Grouped MSE box plots by Forecast Length on Synthesized Datasets, averaged over all sensors trained on \MyTransformer\ and PatchTST. Left to right, the complexity of datasets is increased. Forecast horizons vary from 96 to 720 time steps, corresponding to 1 up to 7 days, assuming the data resolution is the same as the real LausanneCity dataset.}
    \label{fig:6subfigures}
\end{figure}

\subsection{Performance Evaluation on the Synthesized Dataset}
To assess scalability under varied hydrological dynamics, we evaluated both \MyTransformer\ and PatchTST on three synthesized datasets representing distinct urban drainage behaviors on different level of complexity: MeteoSwiss-generated rainfall-driven flows (\textit{SynthLow}) \cite{meteoswiss}, Lorenz Attractor (\textit{SynthMid}) \cite{Lorenz}, and Random Fields (\textit{SynthHigh}) \cite{muller2022gstools} simulating spatially random interactions to analyze the performance of our model in extreme scenarios. To study the scalability of the model, each dataset contains more nodes than the real dataset. 100 nodes randomly selected from the city graph, and performance is averaged across all nodes for the same forecast horizons (96 to 720 time steps, corresponding to 1 up to 7 days). 

Figure~\ref{fig:6subfigures} presents grouped boxplots of MSE distributions under three rainfall configurations: \textit{NoRain}, \textit{RainHist}, and \textit{RainFull} as explained in Table~\ref{table:dataset_version}.
In all data sets and configurations, \MyTransformer\ consistently outperforms PatchTST, with a clear reduction in both median MSE and variance. The performance gap is especially pronounced when forecasted rain is included. 

\begin{figure}
    \centering
    \begin{subfigure}{\linewidth}
        \centering
        \includegraphics[width=0.48\linewidth]{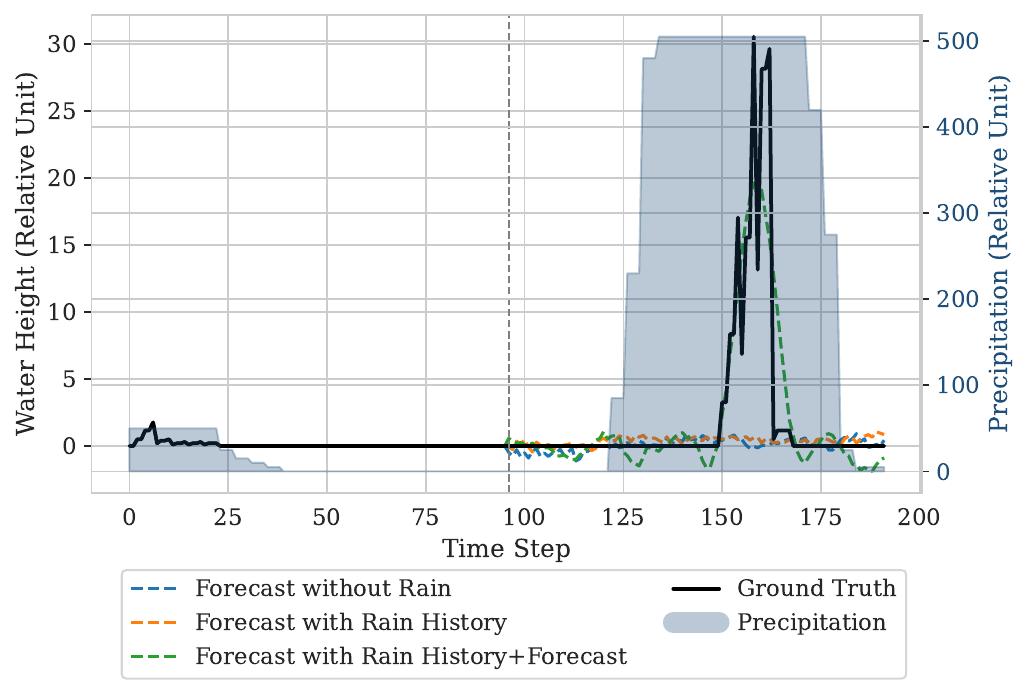}  
        \includegraphics[width=0.48\linewidth]{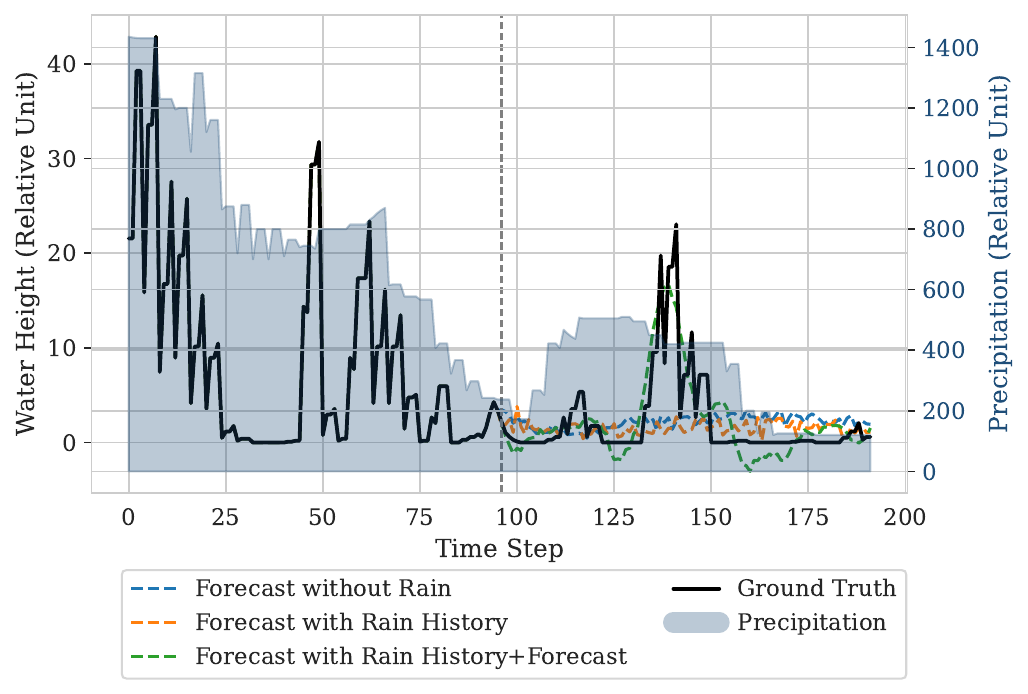}  
    \caption{MeteoSwiss source of cloud for \textit{SynthLow} data configuration}
    \label{fig:forecast-samples-meteo}
    \end{subfigure}

    \centering
    \begin{subfigure}{\linewidth}
        \centering
        \includegraphics[width=0.48\linewidth]{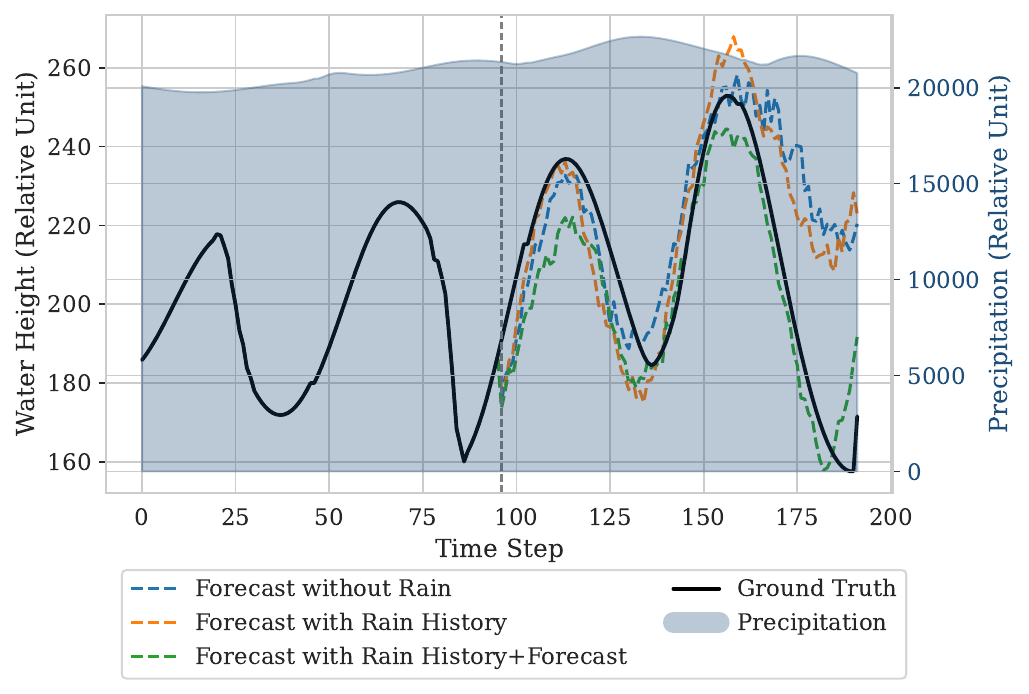}  
        \includegraphics[width=0.48\linewidth]{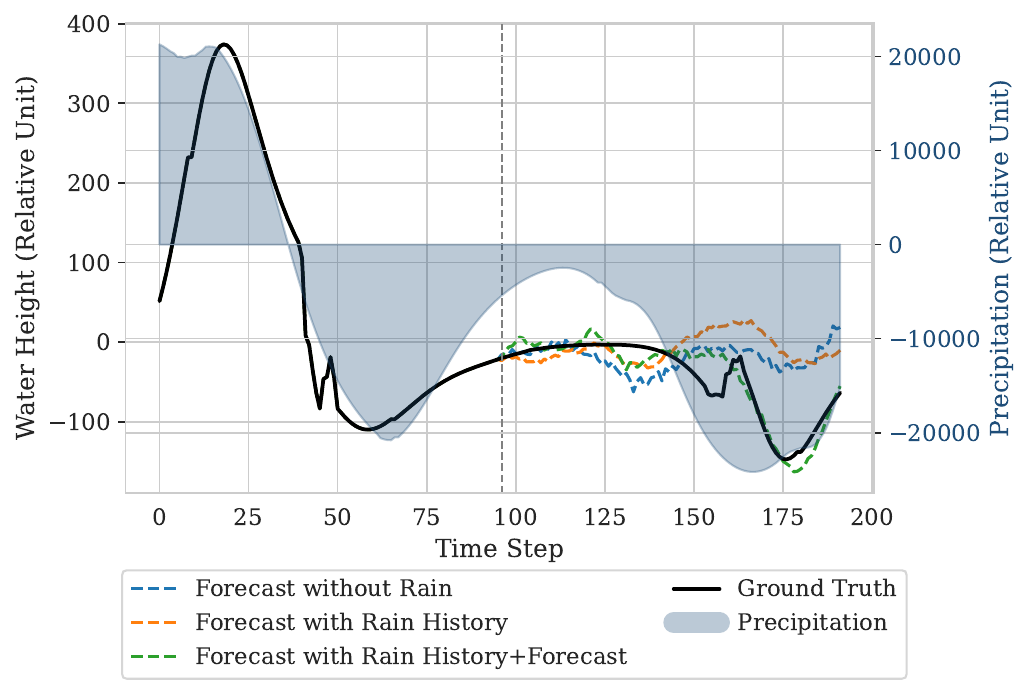}  
        \caption{Lorenz Attractor source of cloud for \textit{SynthMid} data configuration}
        \label{fig:forecast-samples-lorenz}
    \end{subfigure}

    \centering
    \begin{subfigure}{\linewidth}
        \centering
        \includegraphics[width=0.48\linewidth]{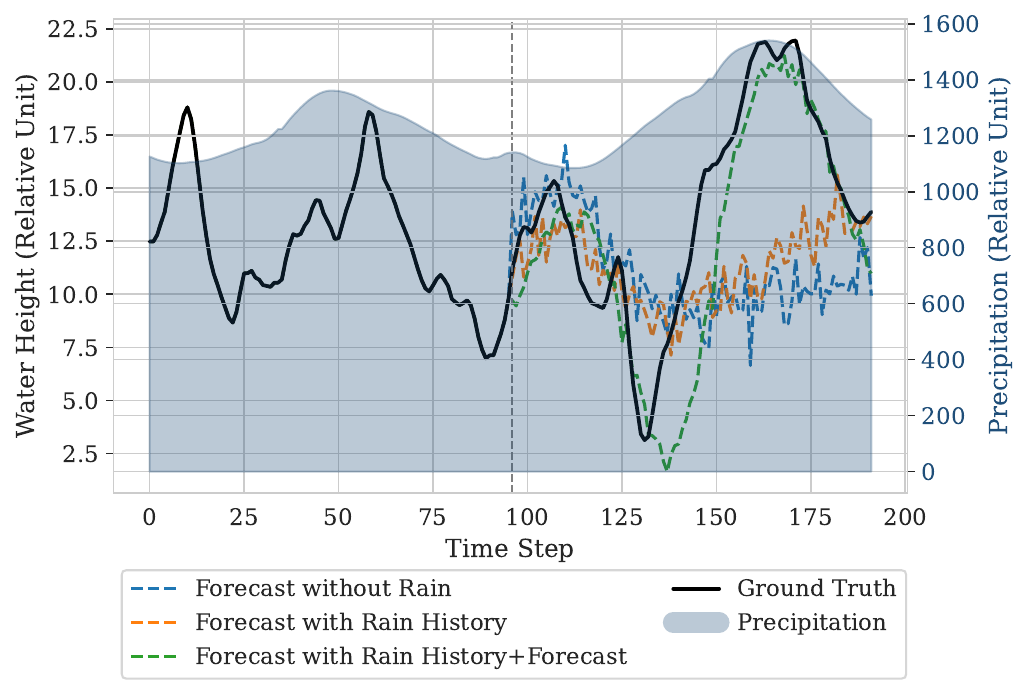} 
        \includegraphics[width=0.48\linewidth]{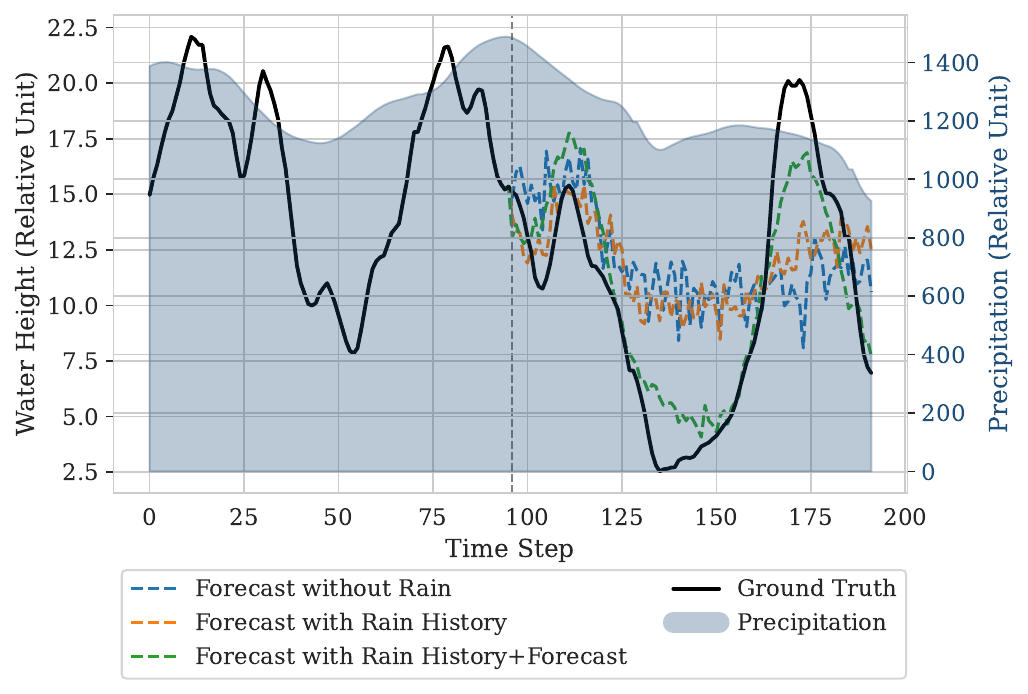}  
        \caption{Random Fields source of cloud for \textit{SynthHigh} data configuration}
    \label{fig:forecast-samples-randomF}
    \end{subfigure}
    
    \caption{\MyTransformer\ forecast samples on Synthesized dataset.}
    \label{fig:forecast-samples-synth}    
\end{figure}

In \textit{SynthLow} dataset, \MyTransformer\ demonstrates significantly improved error as forecast horizon increases, particularly under the \textit{RainFull} dataset. For instance on 720 time-steps equivalent to 7 days forecast, median of MSE decreases from 0.7100 to 0.1200 in Figure~\ref{fig:box-synthL-myT}. Figure~\ref{fig:forecast-samples-meteo} also illustrate two samples, how \textit{RainFull} enable the forecasting to follow the peaks. In contrast, PatchTST shows little to no benefit Figure~\ref{fig:box-synthL-patchtst} from added rain information and maintains relatively flat error distributions across versions, following our results on real dataset in Subsection~\ref{subsec:real}. This reinforces the limitation of PatchTST’s channel-independent design, which cannot exploit cross-variable dependencies.

The \textit{SynthMid} dataset further illustrates the previous contrast. Although the accuracy of PatchTST deteriorates with increasing forecast length and remains largely unaffected by exogenous input, \MyTransformer\ benefits notably from both the rain history and forecasts at short and medium horizons, presented in Figures~\ref{fig:box-synthM-myT} and~\ref{fig:box-synthM-patchtst}. Samples in Figure~\ref{fig:forecast-samples-lorenz} visualize the behavior of \textit{Synth\_mid} and the best performance of \MyTransformer\ on \textit{RainFull} data configuration.
For instance on 1 day history and forecast length (96 time-steps), MSE median of \MyTransformer\ decreases from 0.2369 to 0.0172 by adding rain forecast but PatchTST stays at 0.687.
The designed architecture allows \MyTransformer\ to capture complex, non-linear interactions between rainfall and water dynamics including the sensors' interactions, resulting in tighter error distributions even under chaotic system dynamics.


In the \textit{SynthHigh} dataset, where stochastic nature highly dominates, \MyTransformer\ still maintains superior performance, though the benefit of rain inputs is less dramatic. Figure~\ref{fig:forecast-samples-randomF} presents representative samples that illustrate both the temporal behavior of the dataset and the quality of the forecasts under different rain input conditions.
According to Figures~\ref{fig:box-synthH-patchtst} and~\ref{fig:box-synthH-myT}, median of MSE for PatchTST varies between 0.7626 and 0.8956 in different cases and \MyTransformer\ as expected decrease the error by integrating the rain forecast to 0.1142 from 0.7936 to forecast 1 day with 96 time-steps. However, \MyTransformer\ outperforms PatchTST in all cases, highlighting its scalability and adaptability in diverse regimes of temporal complexity. Results of all metrics reported in Table~\ref{table:synth}. Overall, these results confirm \MyTransformer{}'s capacity to generalize beyond site-specific tuning, making it a strong candidate for large-scale deployment in urban water forecasting systems.

\section{Discussion}
\label{sec:discussion}

This section discusses the key findings in relation to model design, dataset characteristics, and practical implications for urban water forecasting. We highlight the influence of forecast window size, examine the role of time series complexity in evaluating datasets, and explore the scalability of the model in spatially distributed urban settings.

\textbf{Forecast Window Size.} Forecasting window size plays a critical role in model performance and should be selected with respect to the specific application. Although shorter horizons may suffice for real-time control or alert systems, longer horizons are necessary for planning and operational decisions. Across both real and synthesized datasets, we observe a consistent trend: as the forecast horizon increases, the model accuracy decreases. In particular, performance does not degrade linearly in relation to forecast time-steps, reflecting the increasing uncertainty associated with long-term forecasts. This trade-off between horizon length and forecast reliability highlights the importance of aligning window size with the temporal demands of the target application.

\textbf{Complexity and Data Correspondence.} To compare the temporal structure of real and synthesized datasets, we used an entropy-based complexity metric scaled between 0 and 1, where higher values indicate greater randomness. We observed that increased complexity of time series increased complexity of time series is associated with higher forecast error, confirming that such time series is associated with higher forecast error, confirming that such time-series are harder to predict. Although this metric is informative, additional complexity measures are needed to better align the synthesized data with real-world behavior. Such metrics can improve the design of synthetic datasets, making them more realistic or more effective for testing the scalability and robustness of the model under challenging conditions.

\textbf{Node Selection and Generalization.} Our results show that increasing the number of nodes can be used to evaluate the model’s ability to generalize across spatially distributed urban areas. While the LausanneCity dataset used in this study is limited to four real nodes, the use of synthesized data allows us to involve more nodes into analysis and simulate larger networks. Although direct comparisons between real and synthetic datasets are not possible due to differing configurations, this approach enables a scalable assessment of the model. It highlights the potential for extending the forecasting framework to more complex urban settings.

\section{Conclusion}
\label{sec:conc}
In this study, we have proposed \MyTransformer, an attention-based forecasting model designed to incorporate both endogenous water characteristics and exogenous time series, including their predicted values, to improve short-term predictions in urban wastewater systems. Our results show that while endogenous variables capture core cyclic behavior, adding rainfall information, especially short-term forecasts, is crucial for modeling rapid dynamics. \MyTransformer\ performs better than baseline models like PatchTST on both real and synthetic datasets, thanks to its capacity to learn inter-variable relationships and adapt to various temporal setups. The model also scales well to larger and more complex urban networks, enabling a wider application. However, users should be aware of the underlying assumptions when applying such a model to new locations, namely, the need for synchronized and high-quality data, the importance of rainfall as a key driver and the need for model tuning to suit local infrastructure and operational conditions. 

\section*{Acknowledgments}
Thanks to the administration of Lausanne City and its water monitoring service (Ville de Lausanne, Service de l’eau), particularly Mr. Yoann Sadowski, for providing the data and insights about Lausanne for this work, This work is partially supported by the UrbanTwin project financed by the ETH Boards Joint Initiative program in the Strategic Area - Energy, Climate and Sustainable Environment.










\begin{table}
\centering
\scriptsize
\begin{tabular}{lll|ccccc|ccccc}
\hline
\multicolumn{3}{c|}{} & \multicolumn{5}{c|}{\textbf{\MyTransformer\ M2M}} & \multicolumn{5}{c}{\textbf{PatchTST M2M}} \\
\cline{4-13}
\textbf{Source} & \textbf{Forecast} & \textbf{Data Configuration} &
MSE$\downarrow$ & MAE$\downarrow$ & RMSE$\downarrow$ & R\textsuperscript{2}$\uparrow$ & AUC$\uparrow$ &
MSE$\downarrow$ & MAE$\downarrow$ & RMSE$\downarrow$ & R\textsuperscript{2}$\uparrow$ & AUC$\uparrow$ \\
\hline
\multirow{12}{*}{\rotatebox[origin=c]{90}{MeteoSwiss}}
 & \multirow{3}{*}{96}  & SynthLow\_NoRain   & 0.5598 & 0.4103 & 0.7464 & 0.3237 & 0.8419 & 0.6313 & 0.3688 & 0.793 & 0.2366 & 0.8281 \\
 &                      & SynthLow\_RainHist   & 0.6157 & 0.4333 & 0.7835 & 0.2588 & 0.8474 & 0.6311 & 0.3693 & 0.7928 & 0.2372 & 0.8278 \\
 &                      & SynthLow\_RainFull   & 0.0489 & 0.0983 & 0.1962 & 0.9374 & 0.9004 & 0.6297 & 0.3694 & 0.7919 & 0.2382 & 0.8274 \\
 \cline{2-13}
 & \multirow{3}{*}{192} & SynthLow\_NoRain   & 0.5947 & 0.4243 & 0.7701 & 0.2281 & 0.8607 & 0.7421 & 0.4171 & 0.8607 & 0.044 & 0.8117 \\
 &                      & SynthLow\_RainHist   & 0.6186 & 0.4327 & 0.7857 & 0.1983 & 0.8601 & 0.7416 & 0.4173 & 0.8604 & 0.0451 & 0.8116 \\
 &                      & SynthLow\_RainFull   & 0.0502 & 0.0997 & 0.1940 & 0.9313 & 0.8974 & & & NA \\
 \cline{2-13}
 & \multirow{3}{*}{480} & SynthLow\_NoRain   & 0.6673 & 0.4510 & 0.8164 & 0.0695 & 0.8785 & 0.8355 & 0.4416 & 0.9118 & -0.1579 & 0.8026 \\
 &                      & SynthLow\_RainHist   & 0.6602 & 0.4479 & 0.8120 & 0.0793 & 0.8788 & 0.8368 & 0.4426 & 0.9126 & -0.1592 & 0.8027 \\
 &                      & SynthLow\_RainFull   & 0.0915 & 0.1354 & 0.2624 & 0.8659 & 0.8876 & & & NA \\
 \cline{2-13}
 & \multirow{3}{*}{720} & SynthLow\_NoRain   & 0.6881 & 0.4578 & 0.8288 & 0.0300 & 0.8919 & 0.8876 & 0.464 & 0.9384 & -0.2400 & 0.7958 \\
 &                      & SynthLow\_RainHist   & 0.6785 & 0.4575 & 0.8231 & 0.0434 & 0.8690 & 0.8894 & 0.4654 & 0.9394 & -0.2417 & 0.7952 \\
 &                      & SynthLow\_RainFull   & 0.1964 & 0.2290 & 0.4252 & 0.7093 & 0.8922 & & & NA \\
\hline
\hline
\multirow{12}{*}{\rotatebox[origin=c]{90}{Lorenz Attractor}}
 & \multirow{3}{*}{96}  & SynthMid\_NoRain   & 0.1939 & 0.3133 & 0.4357 & 0.8076 & 0.8776 & 0.6632 & 0.5526 & 0.8125 & 0.3713 & 0.7234 \\
 &                      & SynthMid\_RainHist   & 0.2262 & 0.3264 & 0.4709 & 0.7778 & 0.8744 & 0.6651 & 0.5534 & 0.8136 & 0.3676 & 0.7227 \\
 &                      & SynthMid\_RainFull   & 0.0348 & 0.1225 & 0.1724 & 0.9630 & 0.9498 & 0.6638 & 0.5523 & 0.8129 & 0.3671 & 0.7228 \\
 \cline{2-13}
 & \multirow{3}{*}{192} & SynthMid\_NoRain   & 0.5191 & 0.5285 & 0.7177 & 0.5080 & 0.8168 & 0.977 & 0.7214 & 0.9853 & 0.0928 & 0.6701 \\
 &                      & SynthMid\_RainHist   & 0.5239 & 0.5132 & 0.7203 & 0.5038 & 0.8127 & 0.9818 & 0.7234 & 0.9877 & 0.0856 & 0.6690 \\
 &                      & SynthMid\_RainFull   & 0.0488 & 0.1518 & 0.2107 & 0.9505 & 0.9460 & & & NA \\
 \cline{2-13}
 & \multirow{3}{*}{480} & SynthMid\_NoRain   & 1.0484 & 0.8779 & 1.0220 & 0.0456 & 0.7374 & 1.3443 & 0.9073 & 1.1542 & -0.2153 & 0.6076 \\
 &                      & SynthMid\_RainHist   & 1.0461 & 0.8924 & 1.0210 & 0.0471 & 0.7208 & 1.3438 & 0.9066 & 1.154 & -0.2183 & 0.6074 \\
 &                      & SynthMid\_RainFull   & 0.0987 & 0.2292 & 0.3065 & 0.9057 & 0.9243 & & & NA \\
 \cline{2-13}
 & \multirow{3}{*}{720} & SynthMid\_NoRain   & 1.0923 & 0.9090 & 1.0439 & 0.0099 & 0.7139 & 1.6823 & 1.0365 & 1.2895 & -0.5087 & 0.5608 \\
 &                      & SynthMid\_RainHist    & 1.0280 & 0.8761 & 1.0131 & 0.0668 & 0.7290 & 1.6819 & 1.0362 & 1.2894 & -0.5122 & 0.5611 \\
 &                      & SynthMid\_RainFull   & 0.1486 & 0.2909 & 0.3778 & 0.8606 & 0.9047 & & & NA \\
\hline
\hline
\multirow{12}{*}{\rotatebox[origin=c]{90}{Random Fields}}
 & \multirow{3}{*}{96}  & SynthHigh\_NoRain   & 0.5599 & 0.5860 & 0.7319 & 0.1184 & 0.7349 & 0.7182 & 0.66 & 0.8349 & -0.1545 & 0.7042 \\
 &                      & SynthHigh\_RainHist   & 0.5764 & 0.5893 & 0.7423 & 0.0914 & 0.7457 & 0.7186 & 0.6603 & 0.8353 & -0.1548 & 0.7039 \\
 &                      & SynthHigh\_RainFull   & 0.2727 & 0.3667 & 0.4697 & 0.6208 & 0.8141 & 0.719 & 0.6606 & 0.8356 & -0.1542 & 0.7037 \\
 \cline{2-13}
 & \multirow{3}{*}{192} & SynthHigh\_NoRain   & 0.6428 & 0.6390 & 0.7891 & -0.0060 & 0.7172 & 0.7981 & 0.7156 & 0.8828 & -0.2717 & 0.6915 \\
 &                      & SynthHigh\_RainHist    & 0.6458 & 0.6349 & 0.7911 & -0.0115 & 0.7341 & 0.7992 & 0.716 & 0.8834 & -0.2728 & 0.6916 \\
 &                      & SynthHigh\_RainFull  & 0.2771 & 0.3775 & 0.4774 & 0.6190 & 0.8364 & & & NA \\
 \cline{2-13}
 & \multirow{3}{*}{480} & SynthHigh\_NoRain   & 0.6719 & 0.6584 & 0.8077 & -0.0351 & 0.6836 & 0.8215 & 0.7253 & 0.8951 & -0.2801 & 0.6975 \\
 &                      & SynthHigh\_RainHist   & 0.6673 & 0.6534 & 0.8048 & -0.0278 & 0.6928 & 0.8225 & 0.7258 & 0.8958 & -0.281 & 0.6975 \\
 &                      & SynthHigh\_RainFull   & 0.2809 & 0.3798 & 0.4797 & 0.6200 & 0.8288 & & & NA \\
 \cline{2-13}
 & \multirow{3}{*}{720} & SynthHigh\_NoRain   & 0.7226 & 0.6816 & 0.8397 & -0.1262 & 0.7076 & 0.8168 & 0.7226 & 0.8919 & -0.2677 & 0.6965 \\
 &                      & SynthHigh\_RainHist   & 0.6625 & 0.6544 & 0.8021 & -0.0212 & 0.6704 & 0.8174 & 0.7229 & 0.8923 & -0.2676 & 0.6964 \\
 &                      & SynthHigh\_RainFull   & 0.2969 & 0.3892 & 0.4942 & 0.5950 & 0.8284 & & & NA \\
\hline
\end{tabular}
\caption{Synthesized Data, Comparison of \MyTransformer and PatchTST across Sensors, Dataset Versions, and Forecast Lengths, Multi-to-Multi}
\label{table:synth}
\end{table}

\bibliographystyle{unsrt}
\bibliography{cas-refs}

\end{document}